\documentclass[runningheads]{llncs}
\usepackage{graphicx}
\usepackage{tikz}
\usepackage{comment}
\usepackage{amsmath} %
\usepackage{amssymb}
\usepackage{color}
\usepackage{multirow}
\usepackage[pagebackref=true,breaklinks=true,colorlinks,bookmarks=false, letterpaper=true,]{hyperref}
\usepackage[width=122mm,left=18mm,right=18mm, paperwidth=158mm, height=193mm,top=22mm,bottom=22mm, paperheight=237mm]{geometry}

\usepackage{caption}
\usepackage{subcaption}

\usepackage[accsupp]{axessibility}  %

\usepackage{cleveref}

\usepackage{overpic}
\usepackage{enumitem} %
\usepackage{overpic} %
\usepackage{color}
\usepackage{booktabs}
\usepackage{gensymb}
\usepackage{colortbl}
\usepackage{pifont}%
\usepackage{bm}
\definecolor{turquoise}{cmyk}{0.65,0,0.1,0.3}
\definecolor{purple}{rgb}{0.65,0,0.65}
\definecolor{dark_green}{rgb}{0, 0.5, 0}
\definecolor{orange}{rgb}{0.8, 0.6, 0.2}
\definecolor{red}{rgb}{0.8, 0.2, 0.2}
\definecolor{darkred}{rgb}{0.6, 0.1, 0.05}
\definecolor{blueish}{rgb}{0.0, 0.3, .6}
\definecolor{light_gray}{rgb}{0.7, 0.7, .7}
\definecolor{pink}{rgb}{1, 0, 1}
\definecolor{greyblue}{rgb}{0.25, 0.25, 1}

\DeclareMathOperator*{\argmin}{arg\,min}

\newcommand{\real}{\mathbb{R}}
\newcommand{\waymo}{\emph{Waymo}}
\newcommand{\nuscenes}{\emph{nuScenes}}

\crefname{section}{\S}{\S\S}
\crefname{subsection}{\S}{\S\S}
\crefname{equation}{\text{Eq}}{\text{Eq}}
\crefname{definition}{\text{Dfn.}}{\text{Dfn.}}
\crefname{tab}{\text{Tab.}}{\text{Tab.}}
\crefname{fig}{\text{Fig.}}{\text{Fig.}}
\crefname{table}{\text{Tab.}}{\text{Tab.}}
\crefname{figure}{\text{Fig.}}{\text{Fig.}}

\usepackage{blindtext}

\renewcommand{\paragraph}[1]{\vspace{1em}\noindent\textbf{#1}.}

\newcommand{\lcircle}[1]{{\hspace{0.1em}\tikz\draw[#1,fill=#1] (0,0) circle (.4ex);}}

\makeatletter
\DeclareRobustCommand\onedot{\futurelet\@let@token\@onedot}
\def\@onedot{\ifx\@let@token.\else.\null\fi\xspace}

\def\eg{\emph{e.g}\onedot} 
\def\ie{\emph{i.e}\onedot}

\def\wrt{w.r.t\onedot} 
\def\etal{\emph{et al}\onedot}
\makeatother

\newcommand{\ego}{\mathrm{ego}}
\newcommand{\pos}{\mathrm{pos}}
\newcommand{\geo}{\mathrm{ego}}     %

\newcommand{\obj}{\mathrm{obj}}
\newcommand{\trans}{\mathrm{trans}}
\newcommand{\type}{*}
\newcommand{\static}{\mathrm{static}}
\newcommand{\loss}{\mathcal{L}}

\newcommand{\T}{\mathbf{T}}
\newcommand{\tvec}{\mathbf{t}}
\newcommand{\Flow}{\mathbf{V}}
\newcommand{\s}{\mathbf{s}}

\newcommand{\quat}{\mathbf{q}}
\newcommand{\SEuc}{\mathrm{SE}(3)}
\newcommand{\Feature}{\mathbf{F}}
\newcommand{\feature}{\mathbf{f}}
\newcommand{\agfeature}{\tilde{\mathbf{f}}}
\newcommand{\pillar}{\mathbf{p}}
\newcommand{\point}{\mathbf{x}}
\newcommand{\offset}{\bm{\delta}}
\newcommand{\MLP}{\mathrm{MLP}}
\newcommand{\PN}{\mathrm{PN}}
\newcommand{\FG}{\mathrm{FG}}
\newcommand{\cat}{\mathrm{cat}}
\newcommand{\textoffset}{\mathrm{offset}}
\newcommand{\PC}{\mathbf{X}}
\newcommand{\PCset}{\mathcal{X}}
\newcommand{\bev}{\mathrm{base}}  %
\newcommand{\motion}{\mathrm{motion}}  %

\definecolor{lossred}{rgb}{1.0, 0.01, 0.24}
\definecolor{lossgreen}{rgb}{0.55, 0.71, 0.0}
\definecolor{lossblue}{rgb}{0.0, 0.44, 1.0}
\definecolor{lossyellow}{rgb}{1.0, 0.66, 0.07}
\definecolor{losspurple}{rgb}{0.76, 0.33, 0.76}
\definecolor{tab10orange}{rgb}{1.0, 0.7, 0.0}
\newcommand{\refpaper}[1]{{\cref{#1}}}

\begin{document}
\pagestyle{headings}
\mainmatter
\def\ECCVSubNumber{2787}  %

\title{Dynamic 3D Scene Analysis by\\Point Cloud Accumulation}%

\titlerunning{Dynamic 3D Scene Analysis by Point Cloud Accumulation}
\author{Shengyu Huang\inst{1} \quad Zan Gojcic\inst{2} \quad Jiahui Huang\inst{3} \\ Andreas Wieser\inst{1} \quad Konrad Schindler\inst{1}}
\institute{$^1$ETH Z\"{u}rich \quad $^2$NVIDIA \quad $^3$BRCist}
\authorrunning{S. Huang, Z. Gojcic, J. Huang, A. Wieser, K. Schinder}

\maketitle

\begin{figure}[th!]
    \centering
    \vspace{-6mm}
    \includegraphics[width=0.9\linewidth]{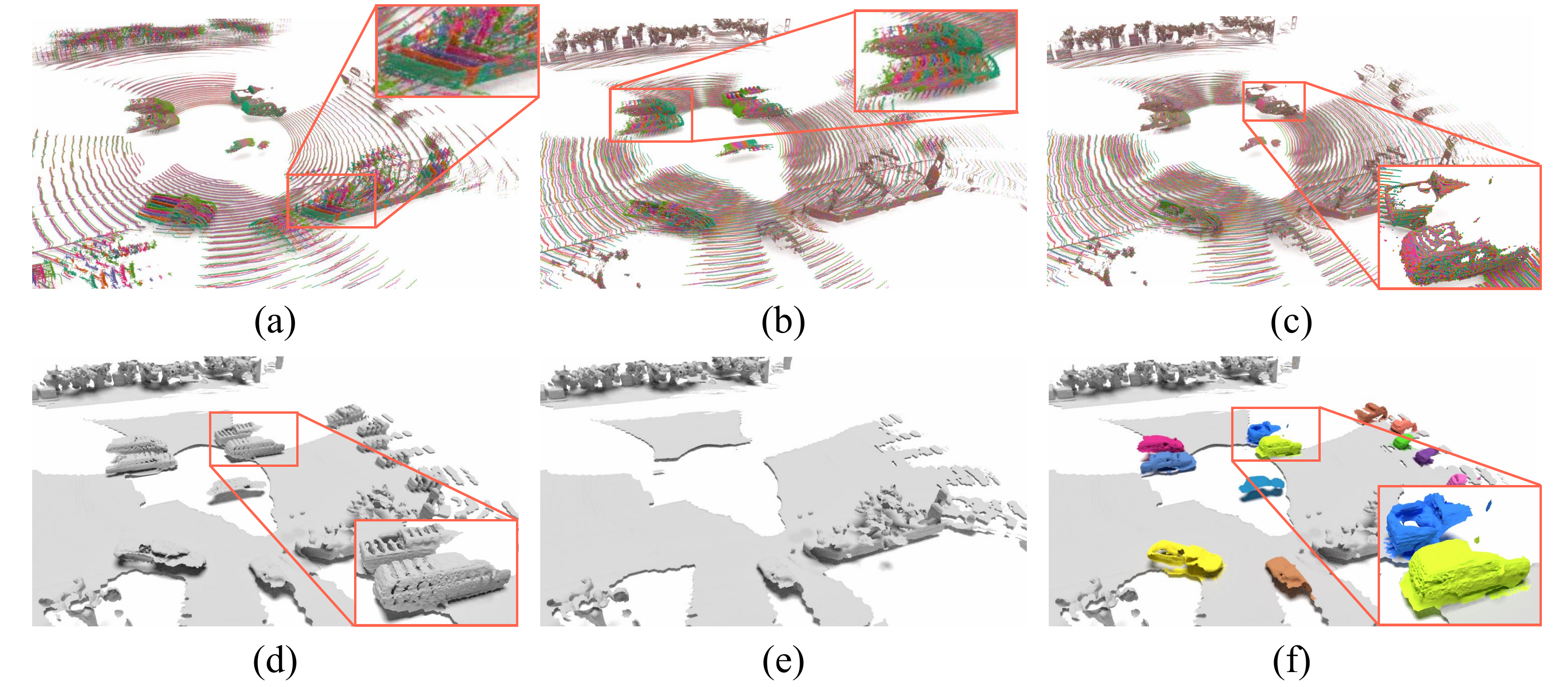}
    \vspace{-0.5em}
    \caption{Points in LiDAR frames acquired over time are not aligned due to the motion of the sensor and of other agents in the scene (a, d). Static background points can be aligned using ego-motion, but this smears the dynamic points across their trajectories (b). While motion segmentation only enables removing the moving points (e), our method properly disentangles individual moving objects from the static part and accumulates both correctly (c, f).}
    \vspace{-10mm}
   \label{fig:teaser}
\end{figure}

\begin{abstract}
Multi-beam LiDAR sensors, as used on autonomous vehicles and mobile robots, acquire sequences of 3D range scans (``frames''). Each frame covers the scene sparsely, due to limited angular scanning resolution and occlusion. The sparsity restricts the performance of downstream processes like semantic segmentation or surface reconstruction. Luckily, when the sensor moves, frames are captured from a sequence of different viewpoints. This provides complementary information and, when accumulated in a common scene coordinate frame, yields a denser sampling and a more complete coverage of the underlying 3D scene. However, often the scanned scenes contain moving objects. Points on those objects are not correctly aligned by just undoing the scanner's ego-motion. In the present paper, we explore multi-frame point cloud accumulation as a mid-level representation of 3D scan sequences, and develop a method that exploits inductive biases of outdoor street scenes, including their geometric layout and object-level rigidity. Compared to state-of-the-art scene flow estimators, our proposed approach aims to align all 3D points in a common reference frame correctly accumulating the points on the individual objects. Our approach greatly reduces the alignment errors on several benchmark datasets. Moreover, the accumulated point clouds benefit high-level tasks like surface reconstruction. [\href{shengyuh.github.io/eccv22/index.html}{\texttt{project page}}]  
\end{abstract}

\section{Introduction}
\label{sec:intro}
LiDAR point clouds are a primary data source for robot perception in dynamically changing 3D scenes. They play a crucial role in mobile robotics and autonomous driving.
To ensure awareness of a large field of view at any point in time, 3D point measurements are acquired as a sequence of sparse scans that each cover a large field-of-view---typically, the full 360$^\circ$.
In each individual scan \emph{(i)} the point density is low, and \emph{(ii)} some scene parts are occluded. Both issues complicate downstream processing.
One way to mitigate the problem is to assume the sensor ego-motion is known and to align multiple consecutive scans into a common scene coordinate frame, thus accumulating them into a denser and more complete point cloud.
This simple accumulation strategy already boosts performance for perception tasks like object detection~\cite{caesar2020nuscenes} and semantic segmentation~\cite{behley2019semantickitti}, but it also highlights important problems.
First, to obtain sufficiently accurate sensor poses the ego-motion is typically computed in post-processing to enable sensor fusion and loop closures --- meaning 
that it would actually not be available for online perception. Second, compensating the sensor ego-motion 
only aligns scan points of the static background, while moving foreground objects are smeared out along their motion trajectories (\cref{fig:teaser}b). 

To properly accumulate 3D points across multiple frames, one must disentangle the individual moving objects from the static background and reason about their spatio-temporal properties. Since the grouping of the 3D points into moving objects itself depends on their motion, the task becomes a form of multi-frame 3D scene flow estimation. Traditional scene flow methods~\cite{liu2019flownet3d,wu2019pointpwc,puy2020flot} model dynamics in the form of a free-form velocity field from one frame to the next, only constrained by some form of (piece-wise) smoothing~\cite{vogel2013piecewise,vogel20153d,dewan2016rigid}.

While this is a very flexible and general approach, it also has two disadvantages in the context of autonomous driving scenes: \emph{(i)} it ignores the geometric structure of the scene, which normally consists of a dominant, static background and a small number of discrete objects that, at least locally, move rigidly;
\emph{(ii)} it also ignores the temporal structure and only looks at the minimal setup of two frames. Consequently, one risks physically implausible scene flow estimates~\cite{gojcic2021weakly}, and does not benefit from the dense temporal sequence of scans.

Starting from these observations, we propose a novel point cloud accumulation scheme tailored to the autonomous driving setting. To that end, we aim to accumulate point clouds over time while abstracting the scene into a collection of rigidly moving agents~\cite{behl2019pointflownet,gojcic2021weakly,teed2021raft}
and reasoning about each agent's motion on the basis of a longer sequence of frames~\cite{huang2021multibodysync,huang2022multiway}. 
Along with the accumulated point cloud (\cref{fig:teaser}c), our method provides more holistic scene understanding, including foreground/background segmentation, motion segmentation, and per-object parametric motion compensation. As a result, our method can conveniently serve as a common, low-latency preprocessing step for perception tasks including surface reconstruction (\cref{fig:teaser}f) and semantic segmentation~\cite{behley2019semantickitti}.

We carry out extensive evaluations on two autonomous driving datasets \emph{Waymo}~\cite{sun2020scalability} and \emph{nuScenes}~\cite{caesar2020nuscenes}, where our method greatly outperforms prior art. For example, on \emph{Waymo} we reduce the average endpoint error from 12.9$\,$cm to 1.8$\,$cm for the static part and from 23.7$\,$cm to 17.3$\,$cm for the dynamic objects. We observe similar performance gains also on \emph{nuScenes}.

In summary, we present a novel, learnable model for temporal accumulation of 3D point cloud sequences over multiple frames, which disentangles the background from dynamic foreground objects. By decomposing the scene into agents that move rigidly over time, our model is able to learn multi-frame motion and reason about motion in context over longer time sequences. Moreover, our method allows for low-latency processing, as it operates on raw point clouds and requires only their sequence order as further input. It is therefore suitable for use in online scene understanding, for instance as a low-level preprocessor for semantic segmentation or surface reconstruction.
\section{Related work}
\label{sec:related}

\paragraph{Temporal point cloud processing}
Modeling a sequence of point clouds usually starts with estimating accurate correspondences between frames, for which scene flow emerged as a popular representation. Originating from \cite{vedula1999three}, scene flow estimation builds an intuitive and effective dynamic scene representation by computing a flow vector for each source point. While traditional scene flow methods~\cite{wedel2008efficient,vogel20113d,vogel2013piecewise,vogel20153d} leverage motion smoothness as regularizer within their optimization frameworks, modern learning-based methods learn the preference for smooth motions directly from large-scale datasets~\cite{liu2019flownet3d,wu2019pointpwc,puy2020flot,ouyang2021occlusion}. Moreover, manually designed scene priors proved beneficial for structured scenes, for instance supervoxel-based local rigidity constraints~\cite{li2021hcrf}, or object-level shape priors learned in fully supervised~\cite{behl2019pointflownet} or weakly supervised~\cite{gojcic2021weakly} fashion.
Methods like SLIM~\cite{baur2021slim} take a decoupled approach, where they first run motion segmentation before deriving scene flows for each segment separately.
Treating the entire point cloud sequence as 4D data and applying a spatio-temporal backbone~\cite{liu2019meteornet,choy2019Minkowski,fan2020pstnet} demonstrates superior performance and efficiency. Subsequent works enhance such backbones by employing long-range modeling techniques such as Transformers~\cite{vaswani2017attention,fan2021point,yang20213d}, or by coupling downstream tasks like semantic segmentation~\cite{aygun20214d}, object detection~\cite{yang2021auto4d,qi2021offboard} and multi-modal fusion~\cite{piergiovanni20214d}.
While our method employs the representation of multi-frame scene flow, we explicitly model individual dynamic objects, which not only provides a high-level scene decomposition, but also yields markedly higher accuracy.

\paragraph{Motion segmentation} 
Classification of the points into static and dynamic scene parts serves as an essential component in our pipeline.
Conventional geometric approaches either rely on ray casting~\cite{chen2016dynamic,schauer2018peopleremover} over dense terrestrial laser scans to build clean static maps, or on visibility~\cite{pomerleau2014long,kim2020remove} to determine the dynamics of the query point by checking its occlusion state in a dense map.
Removert~\cite{kim2020remove} iteratively recovers falsely removed static points from multi-scale range images. Most recently, learning-based methods formulate and solve the segmentation task in a data-driven way: Chen \etal~\cite{chen2021ral} propose a deep model over multiple range image residuals and show SoTA results on a newly-established motion segmentation benchmark~\cite{behley2019semantickitti}. Any Motion Detector~\cite{filatov2020any} first extracts per-frame features from bird's-eye-view projections and then aggregates temporal information from ego-motion compensated per-frame features (in their case with a with convolutional RNN). Our work is similar in spirit, but additionally leverages information from the foreground segmentation task and object clustering in an end-to-end framework.

\paragraph{Dynamic object reconstruction} 
Given sequential observations of a rigid object, dynamic object reconstruction aims to recover the 3D geometric shape as well as  its rigid pose over time. Such a task can be handled either by directly hallucinating the full shape or by registering and accumulating partial observations. Approaches of the former type usually squash partial observations into a global feature vector~\cite{yuan2018pcn,giancola2019leveraging,li2019pu} and ignore the local geometric structure. \cite{gu2020weakly} go one step further by disentangling shape and pose with a novel supervised loss. However, there is still no guarantee for the fidelity of the completed shape. We instead rely on registration and accumulation. Related works include AlignNet-3D~\cite{gross2019alignnet} that directly regresses the relative transformation matrix from concatenated global features of the two point clouds. NOCS~\cite{wang2019normalized} proposes a category-aware canonical representation that can be used to estimate instance pose \wrt its canonical pose. Caspr~\cite{rempe2020caspr} implicitly accumulates the shapes by mapping a sequence of partial observations to a continuous latent space. \cite{huang2021multibodysync} and \cite{huang2022multiway} respectively propose multi-way registration methods that accumulate multi-body and non-rigid dynamic point clouds, but do not scale well to large scenes.
\section{Method}
\label{sec:method}
\begin{figure}[t!]
     \centering
        \includegraphics[width=1.0\linewidth]{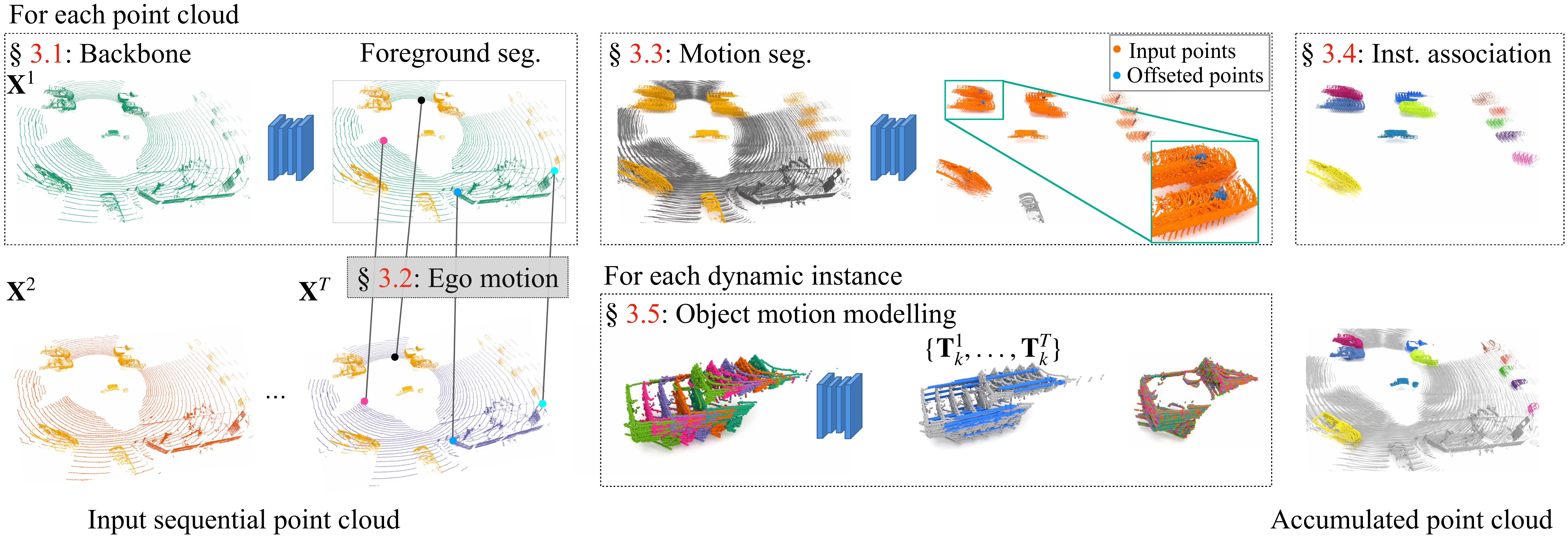}
        \caption{\textbf{Overview}. Our method takes in a point cloud sequence of $T$ frames and starts by extracting foreground points (marked {\setlength{\fboxsep}{0pt}\colorbox{yellow}{yellow}}) for each frame. To obtain ego motion, $T-1$ pairwise registrations are performed. %
       Next, points belonging to dynamic foreground object are extracted using our motion segmentation module (marked {\setlength{\fboxsep}{0pt}\colorbox{tab10orange}{orange}}).
       To boost subsequent spatio-temporal instance association, we additionally predict per-point offset vectors. After instance association, we finally compute the rigid motion separately for each segmented dynamic object.}
 	    \vspace{-6mm}
        \label{fig:architecture}
\end{figure}
The network architecture of our multitask model is schematically depicted in \cref{fig:architecture}. To accumulate the points over time, we make use of the inductive bias that scenes can be decomposed into agents that move as rigid bodies~\cite{gojcic2021weakly}. We start by extracting the latent base features of each individual frame~(\cref{sec:backbone_network}), which we then use as input to the task-specific heads. To estimate the ego-motion, we employ a differentiable registration module (\cref{sec:ego_motion}). Instead of using the ego-motion only to align the static scene parts, we also use it to spatially align the base features, which are reused in the later stages. To explain the motion of the dynamic foreground, we utilize the aligned base features and perform motion segmentation (\cref{sec:motion_seg}) as well as spatio-temporal association of dynamic foreground objects (\cref{sec:clustering}). Finally, we decode the rigid body motion of each foreground object from its spatio-temporal features (\cref{sec:dynamic_obj}). We train the entire model end-to-end with a loss $\loss$ composed of five terms: 
\begin{equation}
    \loss = \loss_\ego^\lcircle{lossred} + \loss_\FG^\lcircle{lossgreen} + \loss_\motion^\lcircle{lossblue} + \lambda_\textoffset\loss_\textoffset^\lcircle{lossyellow} + \lambda_{\text{obj}}\loss_{\text{obj}}^\lcircle{losspurple}\;.
    \label{eq:loss}
\end{equation}

In the following, we provide a high-level description of each module. 
Detailed network architectures, including parameters and loss formulations, are given in the supplementary material (unless already elaborated).

\paragraph{Problem setting}
Consider an \textit{ordered} point cloud sequence $\PCset = \{\PC^t\}_{t=1}^T$ consisting of $T$ frames $\PC^t = [\point_1^t,...,\point_i^t,...,\point_{n_t}^t] \in \mathbb{R}^{3 \times n_t}$ of variable size, captured by a \textit{single moving observer} at constant time intervals $\Delta t$. We denote the first frame $\PC^1$ the \textit{target} frame, while the remaining frames $\{ \PC^{t} \; | \; t>1 \}$ are termed \textit{source} frames. Our goal is then to estimate the flow vectors $\{ \Flow^t \in \mathbb{R}^{3 \times n_t} | t>1 \}$ that align each of the source frames to the target frame, and hence \emph{accumulate} the point clouds. Instead of predicting unconstrained pointwise flow vectors, we make use of the inductive bias that each frame can be decomposed into a %
\textit{static} part $\PC^t_\static$ and $K_{t}$ rigidly-moving \textit{dynamic} parts $\PCset^t_{\text{dynamic}} = \{\PC^t_k\}_{k=1}^{K_t}$~\cite{gojcic2021weakly}. 
For an individual frame, the scene flow vectors $\Flow_\static^t$ of the static part and $\Flow_k^t$ of the $k$-th dynamic object can be explained by the rigid %
ego-motion
$\T_\ego^t \in \SEuc$ and the motion of the dynamic object relative to the static background $\T_k^t \in \SEuc$, respectively as:
\begin{equation}
\label{eq:sf}
    \Flow_\static^t = \T_\ego^t \circ \PC_\static^t - \PC_\static^t, \qquad
    \Flow_k^t = \T_k^t \T_\ego^t \circ \PC_k^t - \PC_k^t,
\end{equation}
where $\T\circ\PC$ (or $\T \circ \point$) denotes applying the transformation to the point set $\PC$ (or point $\point$).

\subsection{Backbone network}
\label{sec:backbone_network}
Similar to~\cite{baur2021slim,jund2021scalable,wu2020motionnet}, our backbone network converts the 3D point cloud of a single frame into a bird's eye view (BEV) 
latent feature image. Specifically, we lift the point coordinates to a higher-dimensional latent space using a point-wise MLP and then scatter them into a $H\times W$ feature grid aligned with the gravity axis. The features per grid cell (``pillar'') are aggregated with max-pooling, then fed through a 2D UNet~\cite{long2015fully} to enlarge their receptive field and strengthen the local context. The output of the backbone network is a 2D latent \emph{base} feature map $\Feature_\bev^t$ for each of the $T$ frames.

% \paragraph{Discussion on point cloud representation}
% An alternative to the BEV projection is sparse voxel~\cite{choy2019Minkowski}. We choose BEV for several reasons: \textit{(i)}. BEV is more efficient in both time and memory, allowing for potential online deployment; \textit{(ii)} BEV allows one to warp, stack, and reuse feature maps in a differentiable manner. Doing the same for sparse voxels is non-trivial; \textit{(iii)} compared to \cite{gojcic2021weakly} that uses a sparse voxel backbone, BEV turns out be to more robust to point density and thus improving the scene flow estimation (see \cref{sec:results}). 

\subsection{Ego-motion estimation}
\label{sec:ego_motion}
We estimate the ego-motion $\T_\ego^t$ using a correspondence-based registration module separately for each source frame. Points belonging to dynamic objects can bias the estimate of ego-motion, especially when using a correspondence-based approach, and should therefore be discarded. 
However, at an early stage of the pipeline it is challenging to reason about scene dynamics, so we rather follow the conservative approach and classify points into background and foreground, where foreground contains all the \emph{movable} objects (\eg, cars and pedestrians), irrespective of their actual dynamics~\cite{gojcic2021weakly}. 
The predicted foreground mask is later used to guide motion segmentation in \cref{sec:motion_seg}.

We start by extracting ego-motion features $\Feature^t_\geo$ and foreground scores $\s^t_\FG$ from each $\Feature_\bev^t$ using two dedicated heads, each consisting of two convolutional layers separated by a ReLU activation and batch normalization. We then randomly sample $N_\ego$ background pillars whose $\s^t_\FG < \tau$ and compute the pillar centroid coordinates $\mathbf{P}^t = \{\pillar_l^t \}$. The ego motion $\T^t_\ego$ is estimated as:
\begin{equation}
{\T^t_\ego} = \argmin_{\T} \sum_{l=1}^{N_\ego} w_l^t \left\|\T \circ \pillar_l^t - \phi(\pillar_l^t, \mathbf{P}^1)\right\|^2. 
\label{eq: ego-motion}
\end{equation}
Here, $\phi(\pillar_l^t, \mathbf{P}^1)$ finds the \emph{soft correspondence} of $\mathbf{p}_l^t$ in $\mathbf{P}^1$, and $w_l^t$ is the weight of the correspondence pair $\big(\pillar_l^t, \phi(\pillar_l^t, \mathbf{P}^1)\big)$. Both $\phi(\pillar_l^t, \mathbf{P}^1)$ and $w_l^t$ are estimated using an entropy-regularized Sinkhorn algorithm from $\Feature_\geo^t$ with slack row/column padded~\cite{cuturi2013sinkhorn,yew2020rpm} and the optimal $\T^t_\ego$ is computed in closed form via the differentiable Kabsch algorithm~\cite{kabsch1976solution}.

We supervise the ego-motion with an $L_1$ loss over the transformed pillars $\loss_\trans =\frac{1}{|\mathbf{P}^t|}  \sum_{l=1}^{|\mathbf{P}^t|} \| \T \circ \pillar_l^t - \overline{\T} \circ \pillar_l^t \|_{1}$ and an inlier loss $\loss_{\text{inlier}}$~\cite{yew2020rpm} that regularizes the Sinkhorn algorithm, $\loss_\ego^\lcircle{lossred}=\loss_\trans + \loss_{\text{inlier}}$.
The foreground score $\s_\FG^t$ is supervised using a combination of weighted binary cross-entropy (BCE) loss $\loss_{\text{bce}}$ and Lovasz-Softmax loss $\loss_{\text{ls}}$~\cite{berman2018lovasz}: $\loss_\FG^\lcircle{lossgreen} = \loss_{\text{bce}}(\s_\FG^t, \overline{\s}_\FG^t) + \loss_{\text{ls}}(\s_\FG^t, \overline{\s}_\FG^t)$, with $\overline{\s}^t_\FG$ the binary ground truth.
The weights in $\loss_{\text{bce}}$ are inversely proportional to the square root of elements in each class.

\subsection{Motion segmentation}
\label{sec:motion_seg}
To separate the \emph{moving} objects from the \emph{static} ones we perform motion segmentation, reusing the per-frame base features $\{\Feature_\bev^t\}$.
Specifically, we apply a differentiable feature warping scheme~\cite{sun2018pwc} that warps each $\Feature_\bev^t$ using the predicted ego-motion $\T^t_\ego$, and obtain a spatio-temporal 3D feature tensor of size {\small${C\times T\times H\times W}$} by stacking the warped feature maps along the channel dimension.
This feature tensor is then fed through a series of 3D convolutional layers, followed by max-pooling across the temporal dimension $T$. 
Finally, we apply a small 2D UNet to obtain the 2D motion feature map $\Feature_\motion$.

To mitigate discretization error, we bilinearly interpolate grid motion features to all foreground points in each frame.\footnote{We predict motion labels only for foreground and treat background points as static.}
The point-level motion feature for $\point_i^t$ is computed as:
\begin{equation}
    \label{eq:point-motion}
    \feature_{\motion,i}^t =\MLP\big(\cat[\psi(\point_i^t,\Feature_\motion), \MLP(\point_i^t)]\big),
\end{equation}
where $\MLP(\cdot)$ denotes a multi-layer perceptron, $\cat[\cdot]$ concatenation, and $\psi(\point, \Feature)$ a bilinear interpolation from $\Feature$ to $\point$.
The dynamic score $\s_i^t$ of the point $\point_i^t$ is then decoded from the motion feature $\feature_{\motion,i}^t$ using another MLP, and supervised similar to the foreground segmentation, with a loss $\loss_\motion^\lcircle{lossblue} = \loss_{\text{bce}}(\s_i^t, \overline{\s}_i^t) + \loss_{\text{ls}}(\s_i^t, \overline{\s}_i^t)$, where $\overline{\s}_i^t$ denotes the ground-truth motion label of point $\point_i^t$.

\subsection{Spatio-temporal instance association}
\label{sec:clustering}
To segment the dynamic points (extracted by thresholding the $\s_i^t$) into individual objects and associate them over time, we perform spatio-temporal instance association.
Different from the common tracking-by-detection~\cite{dendorfer2021motchallenge,weng20203d} paradigm, we propose to directly \emph{cluster the spatio-temporal point cloud}, which simultaneously provides instance masks and the corresponding associations.
However, naive clustering of the ego-motion aligned point clouds often fails due to LiDAR sparsity and fast object motions, hence we predict a per-point offset vector $\offset_i^t$ pointing towards %
the (motion-compensated) instance center:
\begin{equation}
    \offset_{i}^t = \MLP\big(\cat[\psi(\point_i^t, \Feature_\motion), \MLP(\point_i^t)]\big).
\end{equation}

The DBSCAN~\cite{ester1996density} algorithm is subsequently applied over the deformed point set $\{\point_i^t + \offset_i^t \; | \; \forall i, \forall t\}$ to obtain an instance index for each point.
This association scheme is simple yet robust, and can seamlessly handle occlusions and mis-detections.
Similar to 3DIS~\cite{lahoud20193d} we supervise the offset predictions $\offset_i^t$ with both an $L_1$-distance loss and a directional
loss:
\begin{equation}
\loss_\textoffset^\lcircle{lossyellow} =\frac{1}{n}  \sum_{\{i,t\}}^n \left( \left\| \offset_i^t  - \overline{\offset}_i^t \right\|_{1} + 1 - \langle \frac{\offset_i^t}{\| \offset_i^t\|}, \frac{\overline{\offset}_i^t}{\| \overline{\offset}_i^t\|} \rangle \right),
\end{equation}
where $\overline{\offset}$ is the ground truth offset $\mathbf{o}-\point$ from the associated instance centroid $\mathbf{o}$ in the target frame, and $\langle \cdot \rangle$ is the inner product.

\subsection{Dynamic object motion modelling}
\label{sec:dynamic_obj}
Once we have spatio-temporally segmented objects, we must
recover their motions at each frame.
As LiDAR points belonging to a single object are sparse and explicit inter-frame correspondences are hard to find, we take a different approach from the one used in the ego-motion head and construct a novel TubeNet to directly regress the transformations.
Specifically, TubeNet takes $T$ frames of the same instance $\mathbf{X}_k$ as input, and regresses its rigid motion parameters $\T^t_k$ as:
\begin{equation}
    \T^t_k = \MLP \left(\cat[ \agfeature_{\motion}, \agfeature_\ego, \agfeature_{\pos}^t, \agfeature_{\pos}^1] \right),
    \label{eq:tubenet}
\end{equation}
where $\agfeature_\motion$ and $\agfeature_\ego$ are instance-level global features obtained by applying PointNet~\cite{qi2017pointnet} to the respective point-level features of that instance, $\agfeature_{\type} = \PN(\{ \feature_{\type,i}^t \; | \; \point_i^t \in \PC_k^t \})$.
Recall that point-level features $\feature_{\type,i}^t$ are computed from $\Feature_{\type}$ via the interpolation scheme described in \cref{eq:point-motion}.
Here, $\agfeature_\motion$ encodes the overall instance motion while $\agfeature_\ego$ supplements additional geometric cues. The feature
$\agfeature_\pos^t = \PN(\PC_k^t)$ is a summarized encoding over individual frames and provides direct positional information for accurate transformation estimation. 
The transformations are initialised to identity and TubeNet is applied in iterative fashion to regress residual transformations relative to the last iteration, similar to RPMNet~\cite{yew2020rpm}.

For the loss function, we choose to parameterise each $\T$ as an un-normalised quaternion $\quat \in \real^{4}$ and translation vector $\tvec \in \real^3$, and supervise it with:
\begin{equation}
    \loss_\obj^\lcircle{losspurple} = \loss_\trans + \frac{1}{T-1} \sum_{t=2}^T \left( \left\|\overline{\tvec}_k^t - \tvec_k^t \right\|_2 + \lambda \left\|\overline{\quat}_k^t - \frac{\quat_k^t}{{\|\quat_k^t}\|} \right\|_2 \right),
\end{equation}
where $\overline{\tvec}_k^t$ and $\overline{\quat}_k^t$ are the ground truth transformation, and $\lambda$ is a constant weight, set to 50 in our experiments. 
$\loss_\trans$ is the same as in the ego-motion (\cref{sec:ego_motion}).

\subsection{Comparison to related work}
\paragraph{WsRSF~\cite{gojcic2021weakly}} Our proposed method differs from WsRSF in several ways: \textit{(i)} WsRSF is a pair-wise scene flow estimation method, while we can handle multiple frames; \textit{(ii)} unlike WsRSF we perform motion segmentation for a more complete understanding of the scene dynamics; \textit{(iii)} our method outputs instance-level associations, while WsRSF simply connects each instance to the complete foreground of the other point cloud.

\paragraph{MotionNet~\cite{wu2020motionnet}} Similar to our method, MotionNet also deals with sequential point clouds and uses a BEV representation. However, MotionNet \textit{(i)} assumes that ground truth ego-motion is available, while we estimate it within our network; and \textit{(ii)} does not provide object-level understanding, rather it only separates the scene into a static and a dynamic part. 

\subsection{Implementation details}
Our model is implemented in pytorch~\cite{NEURIPS2019_9015} and can be trained on a single RTX 3090 GPU. During training we minimize \cref{eq:loss} with the Adam~\cite{kingma2014adam} optimiser, with an initial learning rate 0.0005 that exponentially decays at a rate of 0.98 per epoch. For both \emph{Waymo} and \emph{nuScenes}, the size of the pillars is $(\delta_x, \delta_y, \delta_z) = (0.25, 0.25, 8)$ m. We sample $N_{\text{ego}} = 1024$ points for ego-motion estimation and set $\tau =0.5$ for foreground/background segmentation. The feature dimensions of $\Feature_\bev^t$, $\Feature_\ego^t$, $\Feature_\motion$ are 32, 64, 64 respectively. During inference we additionally use ICP~\cite{besl1992method} to perform test-time optimisation of the ego-motion as well as the transformation parameters of each dynamic object. ICP thresholds for ego-motion and dynamic object motion are 0.1/0.2 and 0.15/0.25 $\,$m for \emph{Waymo} / \emph{nuScenes}.

\section{Experimental Evaluation}
\label{sec:results}
In this section, we first describe the datasets and the evaluation setting for our experiments (\cref{sec:dataset_evalsetting}). We then proceed with a quantitative evaluation of our method and showcase its applicability to downstream tasks with qualitative results for surface reconstruction~(\cref{sec:evaluation_results}). Finally, we validate our design choices in an ablation study~(\cref{sec:ablation}).

\subsection{Datasets and evaluation setting}
\label{sec:dataset_evalsetting}

\paragraph{Waymo} The Waymo Open Dataset~\cite{sun2020scalability} includes 798/202 scenes for training/validation, where each scene is a 20-second clip captured by a 64-beam LiDAR at 10~Hz. We randomly sample 573/201 scenes for training/validation from the training split, and treat the whole validation split as a held-out test set. We consider every 5 consecutive frames as a \textit{sample} and extract 20 \textit{samples} from each clip, for a total of 11440/4013/4031 samples for training/validation/test. %
    
\paragraph{nuScenes} The nuScenes dataset~\cite{caesar2020nuscenes} consists of 700 training and 150 validation scenes, where each scene is a 20-second clip captured by a 32-beam LiDAR at 20$\,$Hz. We use all validation scenes for testing and randomly hold out 150 training scenes for validation. We consider every 11 consecutive frames as a \textit{sample}, resulting in a total of 10921/2973/2973 samples for training/validation/testing.

\paragraph{Ground truth} We follow~\cite{jund2021scalable} to construct pseudo ground-truth from the detection and tracking annotations. Specifically, the flow vectors of the background part are obtained from ground truth ego-motion. For foreground objects, we use the
unique instance IDs of the tracking annotations and recover their rigid motion parameters by registering the bounding boxes.
Notably, for \nuscenes~the bounding boxes are only annotated every 10 frames. To obtain pseudo ground truth for the remaining frames, we linearly interpolate the boxes, which may introduce a small amount of label noise especially for fast-moving objects.

\paragraph{Metrics}
\begin{table}[t]
    \setlength{\tabcolsep}{4pt}
    \renewcommand{\arraystretch}{1.2}
	\centering
	\resizebox{\columnwidth}{!}{
    \begin{tabular}{clcccc|cccccc}
    \toprule
    & & \multicolumn{4}{c|}{Static part} & \multicolumn{5}{c}{Dynamic foreground}  \\
    Dataset                   & Method    & EPE avg.$\downarrow$ & AccS$\uparrow$  & AccR$\uparrow$ & ROutlier$\downarrow$ & EPE avg. $\downarrow$ & EPE med.$\downarrow$  & AccS$\uparrow$   & AccR$\uparrow$   & ROutliers $\downarrow$                 \\
    \midrule
    \multirow{7}{*}{\emph{Waymo}}    & PPWC-Net~\cite{wu2019pointpwc}  & 0.414 & 17.6 & 40.2 & 12.1 & 0.475 & 0.201 & 9.0 & 29.3 & 22.4  \\
                              & FLOT~\cite{puy2020flot} & 0.129 & 65.2 & 85.0 & 2.8 & 0.625 & 0.231 & 9.8 & 27.4 & 33.8   \\
                              & WsRSF~\cite{gojcic2021weakly} & 0.063 & 87.3 & 96.6 & 0.6 & 0.381 & 0.094 & 31.3 & 64.0 & 10.1       \\
                              & NSFPrior~\cite{li2021neural} & 0.187 & 79.8 & 89.1 & 4.7 & 0.237 & 0.077 & 44.7 & 68.6 & 11.5\\                     \arrayrulecolor{lightgray}\cline{2-11}\arrayrulecolor{black}
                              & Ours & \underline{0.028} & \underline{97.5} & \underline{99.5} & \underline{0.1} & \underline{0.197} & \underline{0.062} & \underline{53.3} & \underline{77.5} & \underline{5.9}  \\
                              & Ours+ & {\bf 0.018} & {\bf 99.0} & {\bf 99.7} & {\bf 0.1} & {\bf 0.173} & {\bf 0.043} & {\bf 69.1} & {\bf 86.9} & {\bf 5.1} \\
                              & Ours (w.\ ground) & 0.042 & 91.9 & 98.8 & 0.1 & 0.219 & 0.071 & 47.1 & 72.8 & 8.5 \\            
    \midrule
    \multirow{7}{*}{\emph{nuScenes}} & PPWC-Net~\cite{wu2019pointpwc}  & 0.316 & 16.1 & 37.0 & 8.7 & 0.661 & 0.307 & 7.6 & 24.2 & 31.9\\
                                  & FLOT~\cite{puy2020flot} & 0.153 & 51.7 & 78.3 & 4.3 & 1.216 & 0.710 & 3.0 & 10.3 & 63.9 \\
                                  & WsRSF~\cite{gojcic2021weakly} & 0.195 & 57.4 & 82.6 & 4.8 & 0.539 & 0.204 & 17.9 & 37.4 & 22.9 \\
                                  & NSFPrior~\cite{li2021neural} &  0.584 & 38.9 & 56.7 & 26.9 & 0.707 & 0.222 & 19.3 & 37.8 & 32.0 \\
                                  \arrayrulecolor{lightgray}\cline{2-11}\arrayrulecolor{black}
                                  & Ours & \underline{0.111} & \underline{65.4} & \underline{88.6} & \underline{1.1} & \underline{0.301} & \underline{0.146} & \underline{26.6} & \underline{53.4} & \textbf{12.1} \\
                                  & Ours+ & {\bf 0.091} & {\bf 72.8} & {\bf 91.9} & {\bf 0.9} & {\bf 0.301} & {\bf 0.135} & {\bf 32.7} & {\bf 56.7} & \underline{13.7} \\
                                  & Ours (w.\ ground) & 0.134 & 55.3 & 83.8 & 1.9 & 0.37 & 0.182 & 18.2 & 43.8 & 17.5  \\   
    \bottomrule
    \end{tabular}
    }
    \vspace{0.5em}
	\caption{Scene flow results on \emph{Waymo} and \emph{nuScenes} datasets.}
	\vspace{-6mm}
	\label{tab:sf_main}
\end{table}
We use standard \emph{scene flow} evaluation metrics~\cite{liu2019flownet3d,baur2021slim} to compare the performance of our approach to the selected baselines. These metric include:
\textit{(i)} 3D end-point-error~(\textit{EPE}~[m]) which denotes the mean $L_2$-error of all flow vectors averaged over all frames; 
\textit{(ii)} strict/relaxed accuracy (\textit{AccS}~[\%]~/\textit{AccR}~[\%]). \ie, the fraction of points with \textit{EPE} $<$ 0.05/0.10m or relative error $<$ 0.05/0.10; \textit{(iii)} \textit{Outliers~[\%]} which denotes the ratio of points with \textit{EPE} $>$ 0.30m or relative error $>$ 0.10; and \textit{(iv)} \textit{ROutliers}~[\%], the fraction of points whose \textit{EPE} $>$ 0.30m and relative error $>$ 0.30. We evaluate these metrics for the static and dynamic parts of the scene separately.\footnote{A point is labelled as \textit{dynamic} if its ground-truth velocity is $>0.5\,$m/s.} %
Following \cite{wu2020motionnet,luo2021self}, we evaluate the performance of all methods only on the points that lie within the square of size $64\times 64$ m$^2$ centered at the ego-car location in the target frame. Additionally we remove ground points by thresholding along the $z$-axis.\footnote{This setting turns out to better suit the baseline methods~\cite{puy2020flot,wu2019pointpwc,li2021neural}. However, we keep ground points in our dynamic point cloud accumulation task, as thresholding could falsely remove points that are of interest for reconstruction or mapping.} Ablations studies additionally report the quality of \textit{motion segmentation} in terms of recall and precision of \textit{dynamic} parts, and the quality of \textit{spatio-temporal instance association} in terms of mean weighted coverage (\textit{mWCov}, the \textit{IoU} of recovered instances~\cite{wang2019associatively}). For further details, see the supplementary material.

\paragraph{Baselines}
We compare our method to 4 baseline methods. PPWC~\cite{wu2019pointpwc} and FLOT~\cite{puy2020flot} are based on dense matching and are trained in a fully supervised manner; WsRSF~\cite{gojcic2021weakly} assumes a \textit{multi-body} scene and can be trained with weak supervision; NSFPrior~\cite{li2021neural} is an optimisation-based method without pre-training. For PPWC~\cite{wu2019pointpwc}, FLOT~\cite{puy2020flot} and WsRSF~\cite{gojcic2021weakly}, we sample at most 8192 points from each frame due to memory constraints, and use the $k$-nn graph to up-sample flow vectors to full resolution at inference time. For NSFPrior~\cite{li2021neural}, we use the full point clouds and take the default hyper-parameter settings given by the authors, except for the early-stopping patience, which we set to 50 to make it computationally tractable on our large-scale dataset. For all baseline methods, we directly estimate flow vectors from any \textit{source} frame to the \textit{target} frame. 

\subsection{Main results}
\label{sec:evaluation_results}
The detailed comparison on the \waymo~and \nuscenes~data is given in \cref{tab:sf_main}. \emph{Ours} denotes the direct output of our model, while \emph{Ours+} describes the results after test-time refinement with ICP. Many downstream tasks (e.g., surface reconstruction) rely on the accumulation of full point clouds and require also ground points. We therefore also train a variant of our method on point clouds with ground points and denote it \emph{Ours (w. ground)}. To facilitate a fair comparison, we use full point clouds as input during training and inference, but only compute the evaluation metrics on points that do not belong to ground. 

\paragraph{Comparison to state of the art}
\begin{figure}[t!]
    \centering
    \includegraphics[width=1.0\linewidth]{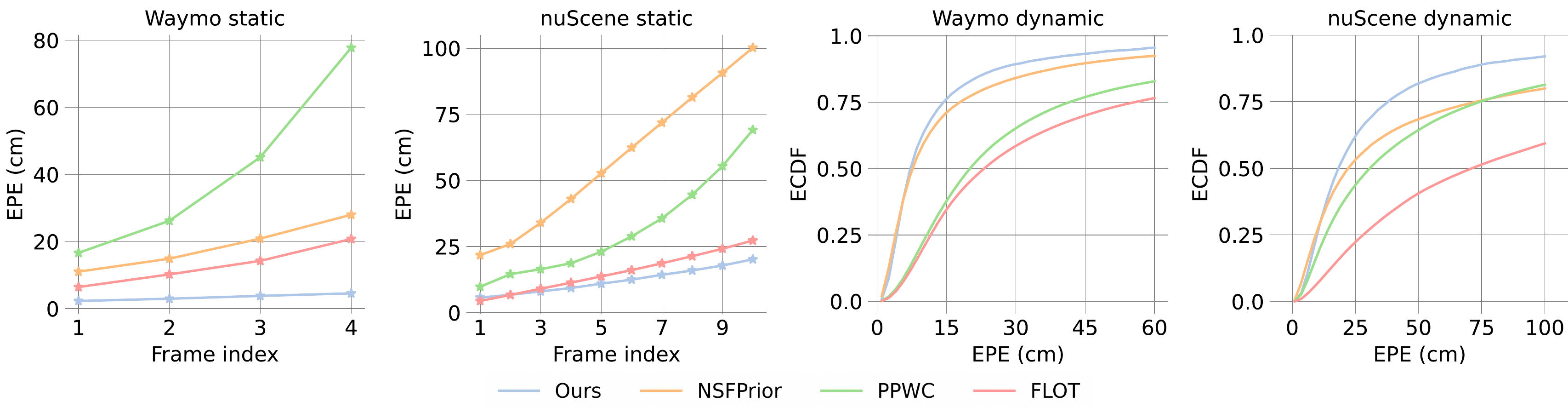}
    \caption{Our method scales better to more frames. For the challenging dynamic parts, it also has smaller errors as well as fewer extreme outliers.}
 	\vspace{-6mm}
    \label{fig:epe3d_ecdf}
\end{figure}
On the static part of the \textbf{\waymo}~dataset, FLOT~\cite{puy2020flot} reaches the best performance among the baselines, but still has an average EPE error of 12.9$\,$cm, which is more than 4 times larger than that of \textit{Ours} (2.8$\,$cm). This result corroborates our motivation for decomposing the scene into a static background and dynamic foreground. Modeling the motion of the background with a single set of transformation parameters also enables us to run ICP at test time (\textit{Ours+}), which further reduces the EPE of the static part to 1.8$\,$cm. On the dynamic foreground points, NSFPrior~\cite{li2021neural} reaches a comparable performance to \textit{Ours}. However, based on our spatio-temporal association of foreground points we can again run ICP at test-time, which reduces the median EPE error to 4.3$\,$cm, $\approx$40$\,\%$ lower than that of NSFPrior. Furthermore, NSFPrior is an optimization method and not amenable to online processing (see Tab.~\ref{tab:runtime}). The results for \textbf{\nuscenes} follow a similar trend as the ones for \emph{Waymo}, but are larger in absolute terms due to the lower point density. Our method achieves the best performance on both static and dynamic parts. The gap to the closest competitors is even larger, which auggests that our method is more robust to low point density.

To further understand the error distribution of the dynamic parts, and the evolution of the errors of the static part as the gap between the \textit{source} and \textit{target} frames increases, we plot detailed results in \cref{fig:epe3d_ecdf}. On both \waymo~and \nuscenes{} our method degrades gracefully, and slower than the baselines. The ECDF curve of the EPE error for foreground points also shows that our method performs best at all thresholds. 

\paragraph{Breakdown of the performance gain}
Overall, our method improves over baseline methods by a large margin on two datasets. The gains are a direct result of our design choices: \textit{(i)}~by modelling the flow of the background as rigid motion instead of unconstrained flow, we can greatly reduce \emph{ROutlier} and improve the accuracy; \textit{(ii)}~different from~\cite{gojcic2021weakly} we perform motion segmentation, and can thus assign ego-motion flow to points on movable, but \emph{static} objects ($\approx75\%$ of the foreground). This further improves the results (see EPE of static FG in \cref{tab:ablations}); \textit{(iii)}~reasoning on the object level, combined with spatio-temporal association and modelling, improves flow estimates for the \textit{dynamic} foreground.

\paragraph{Generalisation to variable input length $T$}
\label{sec: more frames}
When trained with a fixed input length ($T=5$ on \emph{Waymo}), our model is able to generalize to different input lengths (see Tab.~\ref{tab:generalisation}). The performance degrades moderately with increasing $T$, as the motions become larger than seen during training. Also, larger displacements make the correspondence problem inherently harder.
\begin{table}[t]
    \setlength{\tabcolsep}{4pt}
    \renewcommand{\arraystretch}{1.2}
	\centering
	\resizebox{0.7\columnwidth}{!}{
    \begin{tabular}{c|cccccccc}
    \toprule
    & 3 & 4 & 5 & 6 & 7 & 8 & 9 & 10 \\
    \midrule
    static EPE avg. & \textbf{0.022} & 0.025 & 0.028 & 0.032 & 0.037 & 0.044 & 0.054 & 0.066\\
    dynamic EPE avg. & 0.199 & \textbf{0.168} & 0.190 & 0.218 & 0.250 & 0.294 & 0.348 & 0.412 \\
    \bottomrule
    \end{tabular}
    }
    \vspace{0.5em}
    \caption{Scene flow results on \emph{Waymo} dataset w.r.t. input length.}
    \vspace{-6mm}
    \label{tab:generalisation}
\end{table}

\begin{table}[t]
	\centering
	\begin{subfigure}[t]{0.3\linewidth}
	    \vskip 0pt
		\centering
		\resizebox{\textwidth}{!}{%
			\begin{tabular}{ccc}
            \toprule
                            & \emph{Waymo} &\emph{nuScenes} \\
            \midrule
            PPWC-Net~\cite{wu2019pointpwc}       &    \phantom{21}\underline{0.608}       &  \phantom{6}\underline{0.990}   \\
            FLOT~\cite{puy2020flot}         &      \phantom{21}1.028     &  \phantom{6}2.010   \\
            WsRSF~\cite{gojcic2021weakly}            &     \phantom{21}1.252      &    \phantom{6}1.460 \\
            NSFPrior~\cite{li2021neural}            &     212.256      &    63.460 \\
            Ours &      \phantom{21}\textbf{0.174}     &  \phantom{6}\textbf{0.250}  \\
            \bottomrule
            \end{tabular}
		}
	\end{subfigure}
	\hspace{5mm}
	\begin{subfigure}[t]{0.34\linewidth}
		\vskip 0pt
		\centering
		\resizebox{\textwidth}{!}{%
			\begin{tabular}{ccc}
            \toprule
                            & \emph{Waymo} &\emph{nuScenes} \\
            \midrule
            ego-motion estimation & 0.100 & 0.188 \\
            motion segmentation & 0.024 & 0.040 \\
            instance association & 0.036 & 0.009 \\
            TubeNet & 0.014 & 0.013\\
            \bottomrule
            \end{tabular}
		}
	\end{subfigure}
	\vspace{0.5em}
	\caption{Runtimes for the \waymo~and \nuscenes~datasets. All numbers are seconds per 5-frame (\emph{Waymo}), respectively 11-frame (\emph{nuScenes}) \textit{sample}.}
 	\vspace{-6mm}
 	\label{tab:runtime}
\end{table}

\paragraph{Runtime}
We report runtimes for our model and several baseline methods on both datasets in \cref{tab:runtime}~(left). Our method is significantly faster than all baselines under the multi-frame scene flow setting. We also report detailed runtimes of our model for individual steps in \cref{tab:runtime}~(right). As we can see, backbone feature extraction and pairwise registration (\textit{ego-motion estimation}) account for the majority of the runtime, 57.5\% on \waymo~and 75.2\% on \nuscenes. Note that this runtime is calculated over all frames, while under a data streaming setting, we only need to run the first part for a single incoming frame, then re-use the features of the previous frames at later stages, which will greatly reduce runtime: after initialisation, the runtime for every new sample decreases to around 0.094$\,$s for \waymo, respectively 0.079$\,$s on \nuscenes. 

\paragraph{Qualitative results}
\begin{figure}[t]
    \centering
    \includegraphics[width=0.88\linewidth]{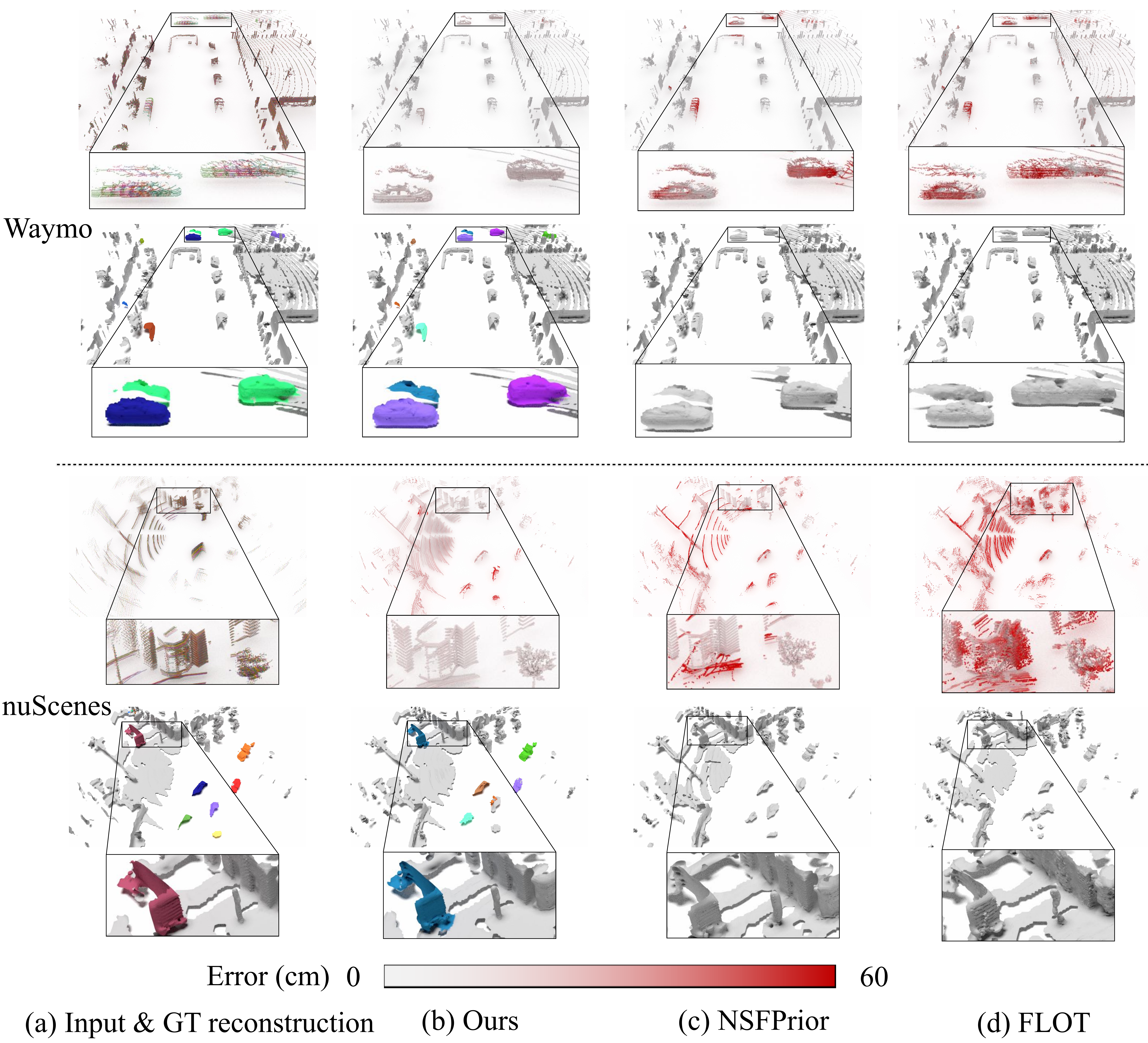}
    \caption{Qualitative results showing scene flow estimation (top) and surface reconstruction (bottom) for two example scenes from \waymo~and \nuscenes.}
    \vspace{-6mm}
    \label{fig:qualitative_results}
\end{figure}
In~\cref{fig:qualitative_results} we show qualitative examples of scene and object reconstruction with our approach. By jointly estimating the ego-motion of the static part and the moving object motions, our method accumulates the corresponding points into a common, motion-compensated frame. It thus provides an excellent basis for 3D surface reconstruction.

\begin{table}[t]
    \setlength{\tabcolsep}{4pt}
    \renewcommand{\arraystretch}{1.2}
    \centering	
    \resizebox{\columnwidth}{!}{
    \begin{tabular}{cccc|cc|c|c|c|ccccc}
    \toprule
      & \multicolumn{3}{c|}{Modules} & \multicolumn{2}{c|}{Motion seg.} & \multicolumn{1}{c|}{Association} &\multicolumn{1}{c|}{Static BG} &  \multicolumn{1}{c|}{Static FG} &  \multicolumn{5}{c}{Dynamic FG} \\
    & FG & MOS & Offset & recall$\uparrow$ & precision$\uparrow$ & WCov$\uparrow$ & EPE avg.$\downarrow$ & EPE avg.$\downarrow$ & EPE avg.$\downarrow$ & EPE med.$\downarrow$ & AccS$\uparrow$ & AccR$\uparrow$ & ROutliers$\downarrow$   \\
    \midrule
    \multirow{4}{*}{\emph{Waymo}} & & \ding{51} & \ding{51} & \underline{92.7} & 94.6 & 82.2 & 0.041 & 0.028 & 0.286 & 0.071 & 45.5 & 71.3 & 10.8\\
    & \ding{51} & & \ding{51} & - & - & \textbf{83.0} & 0.031 & 0.190 & 0.198 & 0.062 & \textbf{53.5} & 77.4 & 6.3\\
    & \ding{51} & \ding{51} & & 92.2 & 96.5 & 79.1 & 0.029 & 0.021 & 0.202 & 0.064  & 52.5 & 76.1 & 6.1\\
    & \ding{51} & \ding{51} & \ding{51} & 92.2 & \textbf{96.8} & 80.4 & \textbf{0.029} & \textbf{0.021} & 0.197 & \textbf{0.062} & 53.3 & \textbf{77.5} & 5.9 \\
    \arrayrulecolor{lightgray}\cline{2-14}\arrayrulecolor{black}
    \multirow{4}{*}{\emph{nuScenes}} & & \ding{51} & \ding{51} & 87.8 & \textbf{92.5} & 65.1 & 0.115 & 0.076 & 0.333 & 0.153 & 25.0 & 50.9 & 13.8\\
    & \ding{51} & & \ding{51} & - & - & \textbf{66.8} & 0.118 & 0.199 & \textbf{0.296} & \textbf{0.143} & \textbf{26.9} & \textbf{53.9} & \textbf{11.7}\\
    & \ding{51} & \ding{51} & & 89.2 & 90.7 & 60.2 & \textbf{0.113} & 0.075 & 0.348 & 0.149 & 25.8 & 51.9 & 13.9 \\
    & \ding{51} & \ding{51} & \ding{51} & \textbf{89.3} & 90.8 & 63.2 & 0.114 & \textbf{0.074} & 0.301 & 0.146 & 26.6 & 53.4 & 12.1\\
    \midrule
    \multirow{2}{*}{\emph{Waymo}} & \multicolumn{3}{c|}{Kalman tracker} & 92.3 & 96.6 & 77.1 & 0.030 & 0.027 & 0.586 & 0.099 & 36.3 & 61.6 & 11.7 \\
    & \multicolumn{3}{c|}{chained poses} & \textbf{93.2} & 94.9 & 81.9 & 0.044 & 0.030 & \textbf{0.171} & 0.068 & 48.0 & 77.0  & \textbf{4.8}\\
    \arrayrulecolor{lightgray}\cline{2-14}\arrayrulecolor{black}
    \multirow{2}{*}{\emph{nuScenes}} & \multicolumn{3}{c|}{Kalman tracker} & 89.4 & 90.8 & 42.9 & 0.114 & 0.092 & 1.238 & 0.364 & 10.7 & 25.1 & 44.0 \\
    & \multicolumn{3}{c|}{chained poses} & 88.1 & 90.2 & 62.1 & 0.225 & 0.151 & 0.315 & 0.155 & 23.4 & 51.8 & 13.4\\
    \bottomrule
    \end{tabular}
    } 
    \vspace{0.5em}
    \caption{Ablation studies on \emph{Waymo} and \emph{nuScenes} datasets.} 
    \vspace{-6mm}
    \label{tab:ablations}
\end{table}
\subsection{Ablation Study}
\label{sec:ablation}
\paragraph{Sequential model}
We evaluate the individual modules of our sequential model, namely the foreground segmentation (\textit{FG}), motion segmentation (\textit{MOS}) and offset compensation (\textit{Offset}). We train two models with and without foreground segmentation. For variants without \textit{MOS} or \textit{Offset}, we take the trained full model but remove \textit{MOS} or \textit{Offset} at inference time. The detailed results are summarised in \cref{tab:ablations}. \textit{FG} enables us to exclude dynamic foreground objects during pairwise registration. On \waymo, this reduces EPE of the static parts by $30\%$ from 4.1 to 2.9$\,$cm, and as a result also reduces EPE of dynamic parts from 28.6 to 19.7$\,$cm. By additionally extracting the static foreground parts with \textit{MOS}, the model can recover more accurate ego-motion for them, which reduces EPE from 19.0/19.9 to 2.1/7.4$\,$cm on \waymo/\nuscenes. \textit{Offset} robustifies the instance association against low point density and fast object motion (+3.1 $pp$ in \textit{WCov} on \nuscenes), this further reduces the EPE of dynamic parts by 3.7$\,$cm. 

\paragraph{Ego-motion estimation strategy}
By default, we directly estimate the ego-motion from any \textit{source} frame to the \textit{target} frame. We compare to an alternative which estimates the ego-motion relative to the previous frame. Although that achieves smaller pairwise errors, after chaining the estimated poses the errors w.r.t.\ the \textit{target} frame explode, resulting in inferior scene flow estimates (\textit{chained poses} in \cref{tab:ablations}).

\paragraph{Comparison to tracking-based method}
Instead of running spatio-temporal association followed by TubeNet to model the motion of each moving object, an alternative would be to apply Kalman tracker so as to simultaneously solve association and motion. We compare to the modified AB3DMOT~\cite{weng20203d}, which is based on a constant velocity model. That method first clusters moving points for each frame independently to obtain instances, then associates instances by greedy matching of instance centroids based on $L_2$ distances. The results in \cref{tab:ablations} (\textit{Kalman tracker}) show clearly weaker performance, due to the less robust proximity metric based on distances between noisy centroid estimates. 
\section{Conclusion}
\label{sec: conclusions}
We have looked at the analysis of 3D point cloud sequences from a fresh viewpoint, as point cloud accumulation across time. In that view we integrate point cloud registration, motion segmentation, instance segmentation, and piece-wise rigid scene flow estimation into a complete multi-frame 4D scene analysis method. By jointly considering sequences of frames, our model is able to disentangle scene dynamics and recover accurate instance-level rigid-body motions. The model processes (ordered) raw point clouds, and can operate online with low latency. A major \emph{limitation} is that our approach is fully supervised and heavily relies on annotated data: it requires instance-level segmentations as well as ground truth motions, although we demonstrate some robustness to label noise from interpolated pseudo ground truth. Also, our system consists of multiple processing stages and cannot fully recover from mistakes in early stages, like incorrect motion segmentation. 

In \emph{future work} we hope to explore our method's potential for downstream scene understanding tasks. We also plan to extend it to an incremental setting, where longer sequences of frames can be summarized into our holistic, dynamic scene representation in an online fashion.

\clearpage
\bibliographystyle{splncs04}

\clearpage
\setcounter{section}{0}
\renewcommand\thesection{\Alph{section}}
In this supplementary document, we first present additional information about our network architecture and implementation in \cref{sec:supp_network}. We then elaborate on technical details of ego-motion estimation, iterative pose refinement, and scene reconstruction in \cref{sec:supp_method}. Precise definitions of loss functions and evaluation metrics are provided in \cref{sec:supp_loss}. Further analysis of the two datasets, \emph{nuScenes} and \emph{Waymo}, is presented in \cref{sec:supp_dataset}, followed by additional quantitative results including scene flow estimation, instance association, and the TubeNet motion model in \cref{sec:supp_quan}. Finally, we show more qualitative results in \cref{sec:supp_qual}. 

\section{Network and implementation}
\label{sec:supp_network}

\begin{figure}[t]
     \centering
        \includegraphics[width=\linewidth]{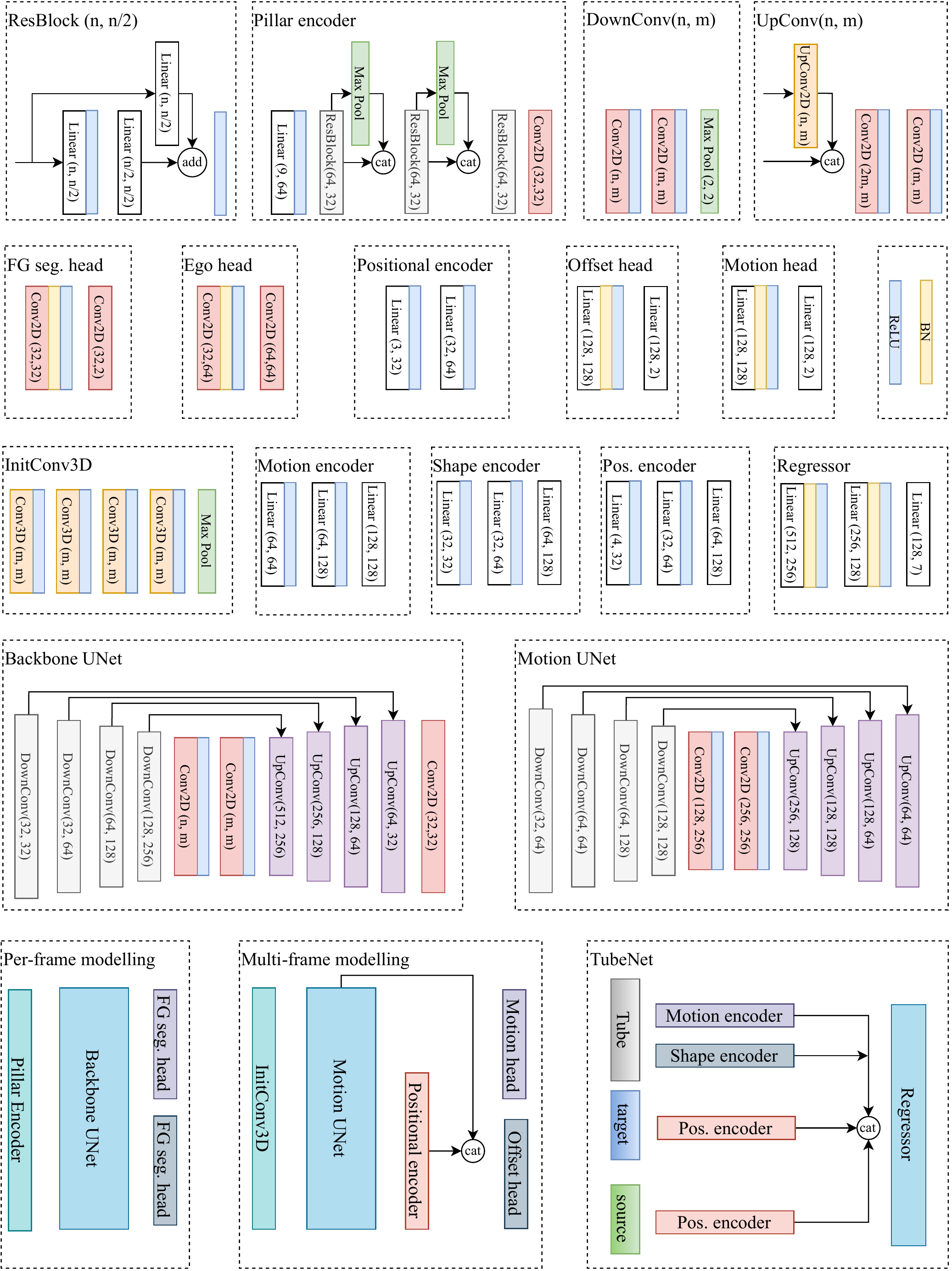}
       \caption{Detailed network architecture. All convolutional layers have kernel size 3$\times$3. $(n, m)$ in Conv, UpConv, DownConv, and Linear layers denote the input and output feature dimensions.} 
   \label{fig:supp_network}
\end{figure}

\paragraph{Network architecture}
The detailed network architecture is depicted in \cref{fig:supp_network}. Our network is a sequential model consisting of (i) per-frame feature extraction used to estimate ego-motion, (ii) multi-frame feature extraction to segment dynamic objects and regress offset vectors towards the associated instance center, and (iii) TubeNet to regress the rigid motions of dynamic objects. The Pillar encoder and the two UNets operate without Batch Normalisation~\cite{ioffe2015batch}, this speeds up training and inference without any loss in performance~\cite{peng2020convolutional}. Our network achieves flexibility w.r.t.\ the number of input frames via global max-pooling along the temporal dimension in the InitConv3D block (\cref{fig:supp_network}).

\paragraph{Implementation details}
We use \texttt{torch\_scatter}\footnote{\url{https://github.com/rusty1s/pytorch_scatter}} to efficiently convert point-wise features to pillar/instance-level global features. To spatially align the backbone features we use \texttt{grid\_sample}, implemented in PyTorch~\cite{NEURIPS2019_9015}. Before clustering, we apply voxel down-sampling implemented in TorchSparse~\cite{tang2022torchsparse} to reduce the point density and improve clustering efficiency. The voxel size is set to 15$\,$cm. Instance labels at full resolution are recovered by indexing points to their associated voxel cell. 

\section{Methodology}
\label{sec:supp_method}
\paragraph{Ego-motion estimation}
Given two sets $(\mathbf{P}^1,\mathbf{P}^t)$ of pillar centroid coordinates and associated $L_2$-normalised features $(\mathbf{F}^1_{\ego}, \mathbf{F}^t_{\ego})$, we first compute the cost matrix $\mathbf{M}^t = 2 - 2\langle \mathbf{F}^t_{\ego}, {\mathbf{F}^1_{\ego}}\rangle$ and an Euclidean distance matrix $\mathbf{D}^t_{l,m} = \|\mathbf{p}^t_l - \mathbf{p}^1_m\|_2$ from pillar coordinates. We then pad $\mathbf{M}^t$ with a learnable slack row and column to accommodate outliers, before iteratively alternating between row normalisation and column normalisation\footnote{To improve training stability, row and column normalisations operate in $\log$-space.} for five times to approximate a doubly stochastic permutation matrix $\mathbf{S}^t$ that satisfies
\begin{equation}
    \sum_{l=1}^{N_\ego+1} \mathbf{S}^t_{l, m} =1, \forall m=1, ..., N_\ego,  \quad \sum_{m=1}^{N_\ego+1} \mathbf{S}^t_{l, m} =1, \forall l=1, ..., N_\ego, \quad \mathbf{S}^t_{l,m} \geq 0. 
\end{equation}
Here $\mathbf{S}^t_{l,m}$ represents the probability of $(\mathbf{p}_l^t, \mathbf{p}_m^1)$ being in correspondence. $\mathbf{p}_l^t$ is considered as an outlier and should be ignored during pose estimation if its slack column value $\mathbf{S}^t_{l,-1} \rightarrow 1$.
We further mask $\mathbf{S}^t$ using a support matrix $\mathbf{I}^t$ computed from $\mathbf{D}^t$ as:
\begin{equation}
    \mathbf{I}^t = \left(\mathbf{D}^t < s\right), \quad s = v \cdot \Delta t\;,
\end{equation}
where $v$ is the maximum speed and $\Delta t$ is the interval between two frames. The final corresponding point $\phi(\mathbf{p}_l^t, \mathbf{P}^1)$ of $\mathbf{p}_l^t$ and its weight $w_l^t$ are computed as
\begin{equation}
    \phi(\mathbf{p}_l^t, \mathbf{P}^1) = (\mathbf{I}^t \odot \mathbf{D}^t)_{[l, :-1]} \PC^1, \quad w_l^t = \sum_{m=1}^{N_\ego} (\mathbf{I}^t \odot \mathbf{D}^t)_{l, m}\;,
\end{equation}
with $\odot$ the Hadamard product. \refpaper{eq: ego-motion} is solved with the Kabsch algorithm. For a detailed derivation, please refer to \cite{gojcic2021weakly}. The value of $v$ is dataset-specific, we set it to 30$\,$m/s for the \emph{Waymo}, respectively 10$\,$m/s for \emph{nuScenes}.

\paragraph{Iterative refinement of TubeNet estimates}
To improve the estimation of the transformation parameters for dynamic objects, we unroll TubeNet for two iterations, as often done in point cloud registration~\cite{yew2020rpm,gojcic2020multiview}. Specifically, for a dynamic object $\PC_k$, we first estimate the initial rigid transformation $\mathbf{T}_k^{0, t}$ of the $t^\text{th}$ frame $\mathbf{X}_k^t$ following \refpaper{eq:tubenet}. We then obtain the transformed points $\mathbf{X}_k^{t'} = \mathbf{T}_k^{0, t} \circ \mathbf{X}_k^t$. Next, we update the positional feature $\agfeature_\pos^{t'} = \PN(\PC_k^{t'})$ and regress the residual transformation matrix $\mathbf{T}_k^{t, 1}$ again, according to \refpaper{eq:tubenet}. The final transformation is $\mathbf{T}_k^{t, 1}\cdot\mathbf{T}_k^{t, 0}$. For better stability during training, the gradients between the two iterations are detached. We assign higher weight to the latter iteration to improve accuracy. The overall loss $\loss_\obj^\lcircle{losspurple}$ is:
\begin{equation}
    \loss_\obj^\lcircle{losspurple} = 0.7\cdot \loss_\obj^{\lcircle{losspurple},0} +  \loss_\obj^{\lcircle{losspurple},1}
\end{equation}

\paragraph{Scene reconstruction}
To show the benefits of our method for downstream tasks, we use the points accumulated with different methods as a basis for 3D surface reconstruction, see \cref{fig:supp_nuscene,fig:supp_waymo}. Specifically, we use the accumulated points as input to the Poisson reconstruction~\cite{kazhdan2006poisson}, implemented in Open3D~\cite{Zhou2018}. To estimate point cloud normals, the neighborhood radius is set to 0.5$\,$m and the maximum number of neighbors is set to 128. The depth for the Poisson method is set to 10, and the reconstructed meshes are filtered by removing vertices with densities below the 15$^\text{th}$ percentile. 

\section{Loss functions and evaluation metrics}
\label{sec:supp_loss}
\subsection{Loss functions}
\paragraph{Weighted BCE loss}
To compensate for class imbalance, we use a weighted BCE and compute the weights of each class on the fly. Specifically, for a mini-batch with $N_{\text{pos}}$ positive and $N_{\text{neg}}$ negative samples, the associated weights $w_{\text{pos}}$ and $w_{\text{neg}}$ are computed as:
\begin{equation}
    w_{\text{pos}} = \min(\sqrt{\frac{N_{\text{pos}} + N_{\text{neg}}}{N_{\text{pos}}}}, w_{\text{max}}) \qquad 
    w_{\text{neg}} = \min(\sqrt{\frac{N_{\text{pos}} + N_{\text{neg}}}{N_{\text{neg}}}}, w_{\text{max}})\;,
\end{equation}
where $w_{\text{max}}$ is the maximum weight of a class.\footnote{We find that in some extreme cases, there are very few ($<10$) positive samples and way more negative samples. We thus bound $w_{\text{max}}$ at 50 to ensure stability.} The final weighted BCE loss $\mathcal{L}_{\mathrm{bce}}(\mathbf{x}, \overline{\mathbf{x}})$ is 
\begin{equation}
\mathcal{L}_{\mathrm{bce}}(\mathbf{x}, \overline{\mathbf{x}}) \!= \frac{1}{|\mathbf{x}|} \sum_{i=1}^{|\mathbf{x}|} w_i (\overline{\mathbf{x}}_i \log(\mathbf{x}_i) + (1 - \overline{\mathbf{x}}_i) \log (1 - \mathbf{x}_i))\;,
\end{equation}
with $\mathbf{x}$ and $\overline{\mathbf{x}}$ are predicted and ground truth labels, and $w_i$ the weight of the $i^\text{th}$ sample, computed as
\begin{equation}
    w_i =      
    \begin{cases}
      w_{\text{pos.}}, & \text{if } \overline{\mathbf{x}}_i = 1  \\
      w_{\text{neg.}}, & \text{otherwise}
    \end{cases}\;.
\end{equation}

\paragraph{Lov\'{a}sz-Softmax loss}
The Jaccard index (ratio of Intersection over Union) is commonly used to measure segmentation quality. In the binary classification setting, we can set the ground truth labels as $\overline{\mathbf{x}}_i \in \{-1,1\}$, then the Jaccard index of the foreground class $J_1$ is computed as
\begin{equation}
    J_1(\overline{\mathbf{x}}, \mathbf{x}) = \frac{|\{\overline{\mathbf{x}} = 1\} \cap \{\mathrm{sign}(\mathbf{x}) = 1\}|}{|\{\overline{\mathbf{x}} = 1\} \cup \{\mathrm{sign}(\mathbf{x}) = 1\}|}\;, \quad J_1(\overline{\mathbf{x}}, \mathbf{x}) \in [0,1]\;,
\end{equation}
with $\mathbf{x}$ the prediction and $\mathrm{sign}()$ the sign function. The corresponding loss $\Delta_{J_1}(\overline{\mathbf{x}}, \mathbf{x})$ to minimise the empirical risk is
\begin{equation}
    \Delta_{J_1}(\overline{\mathbf{x}}, \mathbf{x}) = 1 - J_1(\overline{\mathbf{x}}, \mathbf{x})\;.
\end{equation}
However, this is not differentiable and cannot be directly employed as a loss function. The authors of~\cite{berman2018lovasz} have proposed to optimise it using a Lov\'{a}sz extension. The Lov\'{a}sz extension $\dddot{\Delta}$ of a set function $\Delta$ is defined as:
\begin{equation}
    \dddot{\Delta} (\mathbf{m}) = \sum_{i=1}^{p} \mathbf{m}_i \, g_i(\mathbf{m})\;,
\end{equation}
with 
\begin{equation}
g_i(\mathbf{m}) =
\Delta(\{\pi_1, \ldots, \pi_i\})- \Delta(\{\pi_1, \ldots, \pi_{i-1}\}),
\end{equation}
where $\mathbf{\pi}$ denotes a permutation that places the components of $\mathbf{m}$ in decreasing order. Considering $\mathbf{m}_i = \max(1 - \mathbf{x}_i \overline{\mathbf{x}}_i, 0)$, the Lov\'{a}sz-Softmax loss $\mathcal{L}_{ls}(\overline{\mathbf{x}}, \mathbf{x})$ is defined as 
\begin{equation}
    \mathcal{L}_{ls}(\overline{\mathbf{x}}, \mathbf{x}) = \dddot{\Delta_{J_1}}(\mathbf{m}).
\end{equation}

\paragraph{Inlier loss}
Previous works~\cite{yew2020rpm,gojcic2021weakly} have observed that the entropy-regularized optimal transport~\cite{cuturi2013sinkhorn} has a tendency to label most points as outliers. To alleviate this issue, we follow~\cite{yew2020rpm,gojcic2021weakly} and use an inlier loss $\mathcal{L}_{\text{inlier}}$ on the matching matrix $\mathbf{D}^t$, designed to encourage inliers. The inlier loss is defined as
\begin{equation}
    \mathcal{L}_{\text{inlier}}^t = \frac{1}{2N_{\ego}} (2N_{\ego} - \sum_{l=1}^{N_\ego} \sum_{m=1}^{N_\ego} \mathbf{D}_{l,m}^t)\;.
\end{equation}

\subsection{Evaluation metrics}
\paragraph{Instance association metrics}
To quantitatively measure the spatio-temporal instance association quality, we report weighted coverage (\textit{WCov}) as well as recall and precision at a certain threshold. Given the ground truth clusters $\mathcal{G}$ and the estimated clusters $\mathcal{O}$, recall measures the ratio of clusters in $\mathcal{G}$ that have an overlap above some threshold with a cluster in $\mathcal{O}$, while precision does the same in the opposite direction. Weighted coverage $WCov (\mathcal{G},\mathcal{O})$ is computed as
\begin{equation}
    WCov(\mathcal{G}, \mathcal{O}) = \sum_{i=1}^{|\mathcal{G}|} \frac{1}{|\mathcal{G}|} w_i \max_j \mathrm{IoU} (r_i^G, r_j^0), \qquad  w_i = \frac{|r_i^G|}{\sum_k |r_k^G|}\;,
\end{equation}
where $r_i^G$ and $r_j^O$ are clusters from $\mathcal{G}$ and $\mathcal{O}$, and $\mathrm{IoU}(r_i^G, r_j^O)$ denotes the overlap between two clusters. 

\paragraph{ECDF}
The Empirical Cumulative Distribution Function (ECDF) measures the distribution of a set of values:
\begin{equation}
\begin{aligned}
\text{ECDF} (x) = \frac{\big|\{o_i < x\}\big|}{\big|O\big|}\;,
\end{aligned}
\end{equation}
where $O = \{o_i\}$ is a set of samples and $x \in [\min\{O\}, \max\{O\}]$.

\section{Dataset analysis}
\label{sec:supp_dataset}
\begin{figure}[t!]
     \centering
        \includegraphics[width=1.0\linewidth]{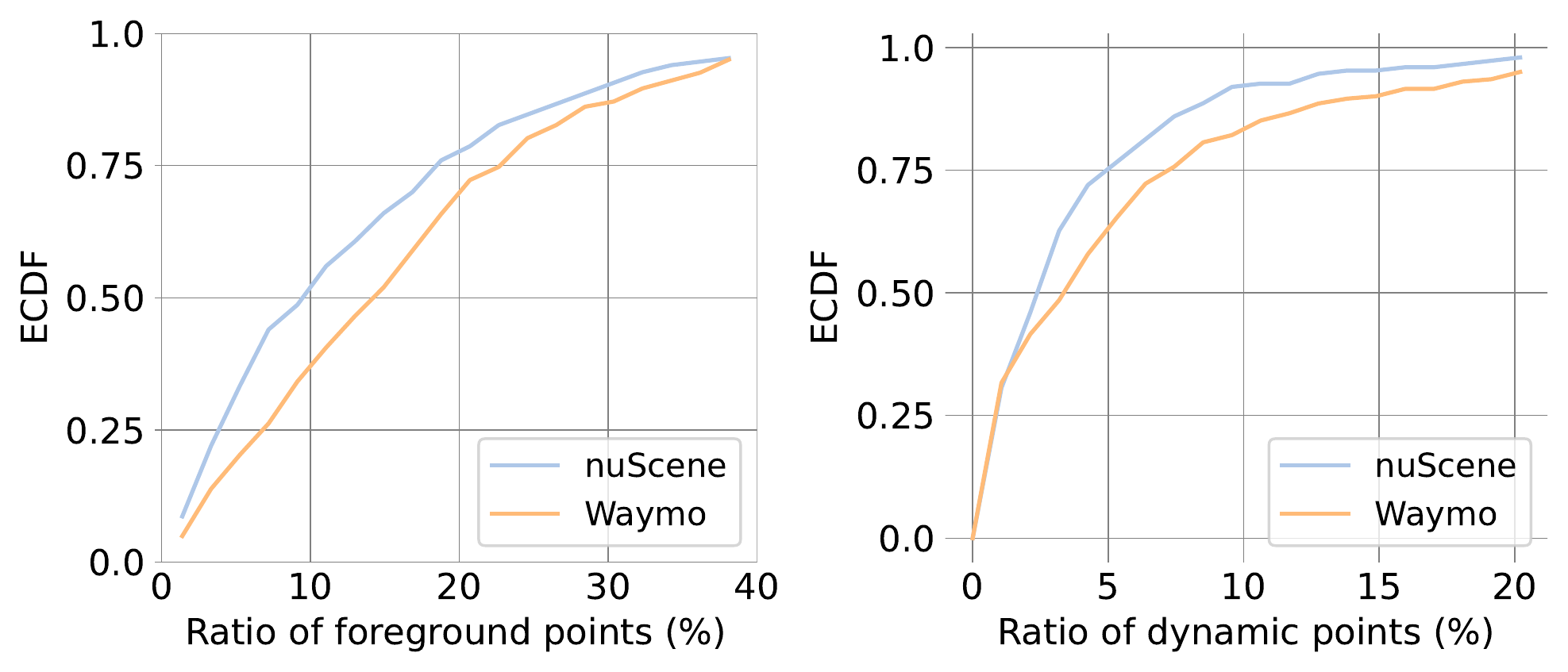}
   \vspace{-0.7cm}
       \caption{\textit{ECDF} curve of points lying on foreground objects and on dynamic objects, for both the \emph{Waymo} and \emph{nuScenes} datasets.}
   \label{fig:dataset_ecdf}
\end{figure}
\begin{figure}[t!]
     \centering
        \includegraphics[width=1.0\linewidth]{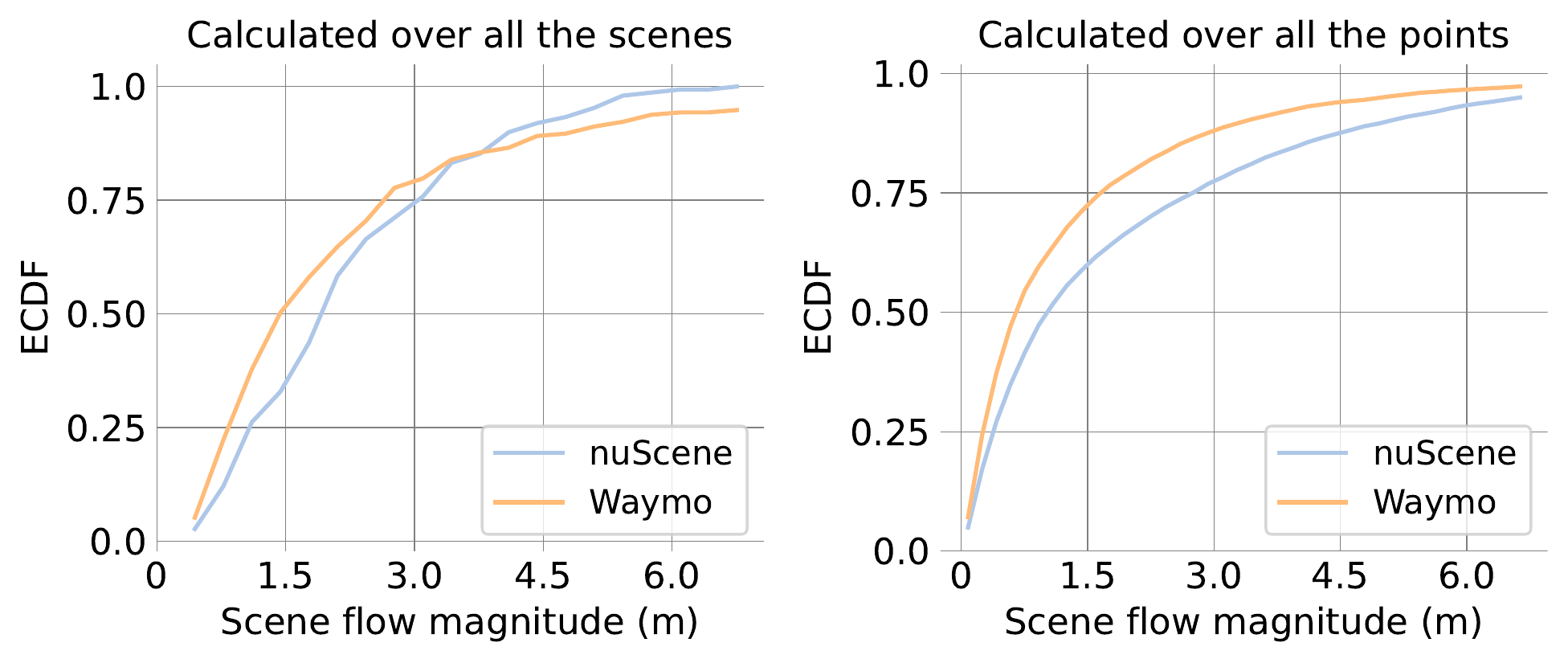}
   \vspace{-0.7cm}
       \caption{Scene flow magnitudes of dynamic objects in the \emph{Waymo} and \emph{nuScenes} datasets.}
   \label{fig:motion_ecdf}
\end{figure}
In total, we have 150, respectiverly 202 scenes as held-out test sets in \emph{nuScenes} and \emph{Waymo}. The \textit{ECDF} curve of points belonging to foreground and dynamic objects are shown in \cref{fig:dataset_ecdf}. As can be seen, the ratios of foreground and dynamic points span a large range (40\% and 20\%). Recalling that the scene flow estimation performance of the dynamic parts falls far behind that of the static parts (\refpaper{tab:sf_main}), this large range of ratios of dynamic objects hints at different difficulties across the scenes. The median fractions of foreground points are 16.2\%/9.4\% in \emph{Waymo}/\emph{nuScenes}, the median fractions of points on moving objects are 3.5\%/2.4\%. In other words, roughly 75\% of all foreground objects are static. This motivates
our strategy to start with motion segmentation, so as to make explicit the large static component (including many objects that could move) whose scene flow is identical to the ego-motion.

In \cref{fig:motion_ecdf}, we show the \textit{ECDF} curve of scene flow magnitudes ($L_2$-norm of scene flow vectors) for the dynamic portions of the two datasets. The motions span a large range, but 75\% of the flow vectors are of moderate magnitude \textless3$\,$m. \emph{nuScenes} has slightly larger overall flow magnitudes than \emph{Waymo}, but \emph{Waymo} contains more instances of large motions (\cref{fig:motion_ecdf}~(left)).

\section{Additional results}
\label{sec:supp_quan}
\paragraph{Results averaged over scenes}
\begin{table}[t]
    \setlength{\tabcolsep}{4pt}
    \renewcommand{\arraystretch}{1.2}
	\centering
	\resizebox{\columnwidth}{!}{
    \begin{tabular}{clcccc|ccccc}
    \toprule
    & & \multicolumn{4}{c|}{Static part} & \multicolumn{4}{c}{Dynamic foreground}  \\
    Dataset  & Method  & EPE avg.$\downarrow$ & AccR$\uparrow$  & AccS$\uparrow$ & ROutlier$\downarrow$ & EPE avg. $\downarrow$ & AccR$\uparrow$   & AccS$\uparrow$   & ROutliers $\downarrow$ \\
    \midrule
    \multirow{5}{*}{\emph{Waymo}} & PPWC-Net~\cite{wu2019pointpwc}  &  0.475 $\pm$ 0.543 & 35.0 & 14.2 & 13.5 & 0.658 $\pm$ 0.696 & 27.1 & 7.9 & 22.9\\
                              & FLOT~\cite{puy2020flot} &   0.381 $\pm$ 0.516 & 68.8 & 51.8 & 13.0 & 0.772 $\pm$ 0.711 & 30.1 & 11.2 & 31.9\\
                              & WsRSF~\cite{gojcic2021weakly} & 1.415 $\pm$ 1.352 & 34.6 & 23.0 & 56.9 & 1.764 $\pm$ 1.744 & 21.0 & 8.6 & 61.6\\
                              & NSFPrior~\cite{li2021neural} & 0.159 $\pm$ 0.231 & 87.1 & 73.5 & 4.3 & 0.355 $\pm$ 0.456 & 63.7 & 41.3 & 14.3 \\
                              & Ours & \bf{0.088} $\pm$ 0.237 & \bf{91.6} & \bf{81.9} & \bf{2.3} & \bf{0.169} $\pm$ 0.259 & \bf{76.8} & \bf{52.9} & \bf{5.3}\\
    \midrule
    \multirow{5}{*}{\emph{nuScenes}} & PPWC-Net~\cite{wu2019pointpwc} & 0.488 $\pm$ 0.402 & 34.2 & 12.7 & 17.5 & 0.784 $\pm$ 0.547 & 22.8 & 6.9 & 35.0 \\
                                  & FLOT~\cite{puy2020flot} & 0.597 $\pm$ 0.582 & 53.3 & 35.1 & 26.6 & 1.156 $\pm$ 0.714 & 13.2 & 3.7 & 56.5\\
                                  & WsRSF~\cite{gojcic2021weakly} & 0.658 $\pm$ 0.483 & 47.5 & 31.1 & 31.5 & 0.925 $\pm$ 0.627 & 29.8 & 15.0 & 42.2\\
                                  & NSFPrior~\cite{li2021neural} & 0.501 $\pm$ 0.344 & 57.8 & 37.7 & 21.3 & 0.743 $\pm$ 0.537 & 39.1 & 19.9 & 31.1 \\
                                   & Ours & \bf{0.226} $\pm$ 0.206 & \bf{72.3} & \bf{46.7} & \bf{7.4} & \bf{0.394} $\pm$ 0.26 & \bf{47.8} & \bf{22.7} & \bf{17.3}\\
    \bottomrule
    \end{tabular}
    }
    \vspace{0.5em}
	\caption{Scene flow estimation results on \emph{Waymo} and \emph{nuScenes} datasets. Numbers are averaged over all test scenes.}
	\vspace{-4mm}
	\label{tab:sf_supp}
\end{table}\textbf{}
In \refpaper{tab:sf_main} we report evaluation metrics calculated over all the points in the test set. However, this does not fully reveal the difficulties encountered in different scenes. Here, we first calculate evaluation metrics per scene, then report the average over scenes in \cref{tab:sf_supp}. For \textit{EPE avg.}, we additionally report the standard deviations. We can see that for both static and dynamic parts, all methods have large standard deviations, which indicates varying difficulty of the scenes, as well as gross errors from challenging samples. Our model still achieves the smallest flow errors and standard deviations under this evaluation setting, for both datasets .

\paragraph{Spatio-temporal instance association}
\begin{figure}[t]
     \centering
        \includegraphics[width=1.0\linewidth]{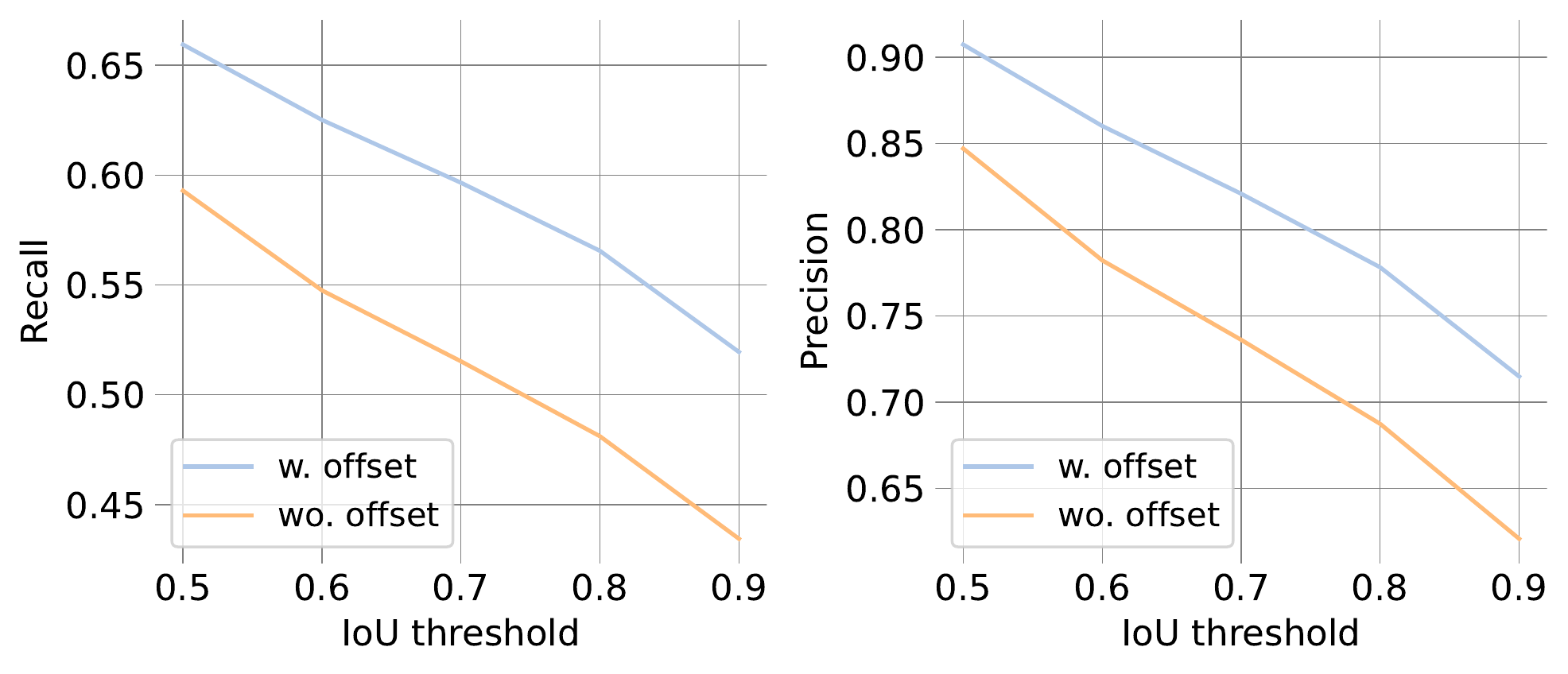}
   \vspace{-0.7cm}
       \caption{Spatio-temporal instance association performance on \emph{Waymo} with and without offset prediction.}
   \label{fig:supp_association}
\end{figure}
We plot instance association metrics at different thresholds in \cref{fig:supp_association}. As can be seen, offset prediction improves association recall and precision by \textgreater5\%, across a range of thresholds. Such improvement becomes more significant as one increases the \textit{IoU} threshold, reaching $\approx$10\% at $IoU=0.9$. We conclude that offset prediction is important to retain high-quality spatio-temporal instances, which can subsequently improve the accuracy of motion modelling for the dynamic parts (\textit{AccS} increases by 9.2\%  in \refpaper{tab:ablations}).

\paragraph{Dynamic object motion modelling}
\begin{table}[t]
    \setlength{\tabcolsep}{4pt}
    \renewcommand{\arraystretch}{1.2}
	\centering
    \begin{tabular}{clcccccc}
    \toprule
   &  & EPE avg.$\downarrow$ & EPE med.$\downarrow$ & AccS$\uparrow$ & AccR$\uparrow$ & ROutliers$\downarrow$   \\
   \midrule
    \multirow{3}{*}{\emph{Waymo}} & center & 0.265 & 0.095 & 36.9 & 62.9 & 9.9\\ 
    & center + ICP  & 0.212 & 0.047 & 61.2 & 80.0 & 7.7 \\
    & Ours   & 0.197 & 0.062 & 53.3 & 77.5 & 5.9 \\
    & Ours + ICP   & \bf{0.173} & \bf{0.043} & \bf{69.1} & \bf{86.9} & \bf{5.1}\\
    \midrule
    \multirow{3}{*}{\emph{nuScenes}} & center & 0.553 & 0.258 & 13.5 & 32.7 & 28.2     \\ 
    & center + ICP   & 0.525 & 0.179 & 23.8 & 43.7 & 25.5\\
    & Ours   & 0.301 & 0.146 & 26.6 & 53.4 & \textbf{12.1}    \\
    & Ours + ICP   & \bf{0.301} & \bf{0.135} & \bf{32.7} & \bf{56.7} & 13.7 \\
    \bottomrule
    \end{tabular}
	\vspace{1mm}
	\caption{Comparison to centroid-based motion estimation baseline.}
	\vspace{-6mm}
	\label{tab:tubenet_comparison}
\end{table}
We additionally compare the proposed TubeNet to two baseline methods. We naively align each frame $\PC_k^t$ $(t>1)$ of an instance $\PC_k$ to frame $\PC_k^0$ by %
translating the centroids, and term this method \textit{center}. For \textit{center+ICP} we refine the simple translational alignment by a subsequent ICP. The detailed comparison is shown in \cref{tab:tubenet_comparison}. Our learned TubeNet achieves the best performance on both datasets. The improvement is larger on the challenging \emph{nuScenes} data, where the point clouds are sparser and less complete, so centroids computed from partial observations are not an accurate proxy for the object location. Our learned TubeNet can implicitly exploit prior knowledge about object shape and surface-level correspondence, leading to more robust and accurate motion modelling. 

\section{Qualitative results}
\label{sec:supp_qual}
\begin{figure}[t]
     \centering
        \includegraphics[width=\linewidth]{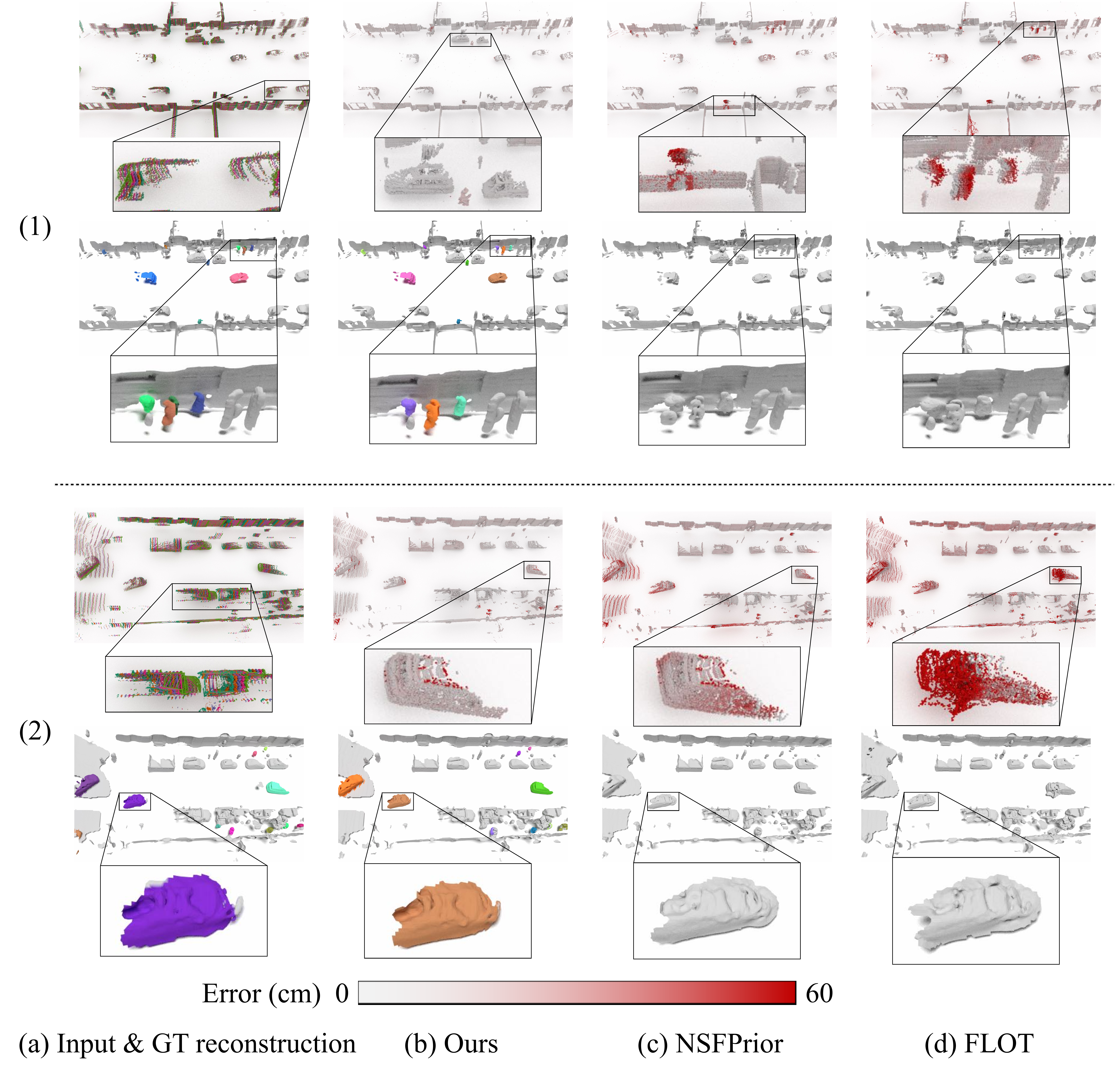}
   \vspace{-1em}
       \caption{Qualitative results showing scene flow estimation (top) and surface reconstruction (bottom) for three example scenes from the \emph{Waymo} dataset.}
   \label{fig:supp_waymo}
\end{figure}

\begin{figure}[t]
     \centering
        \includegraphics[width=\linewidth]{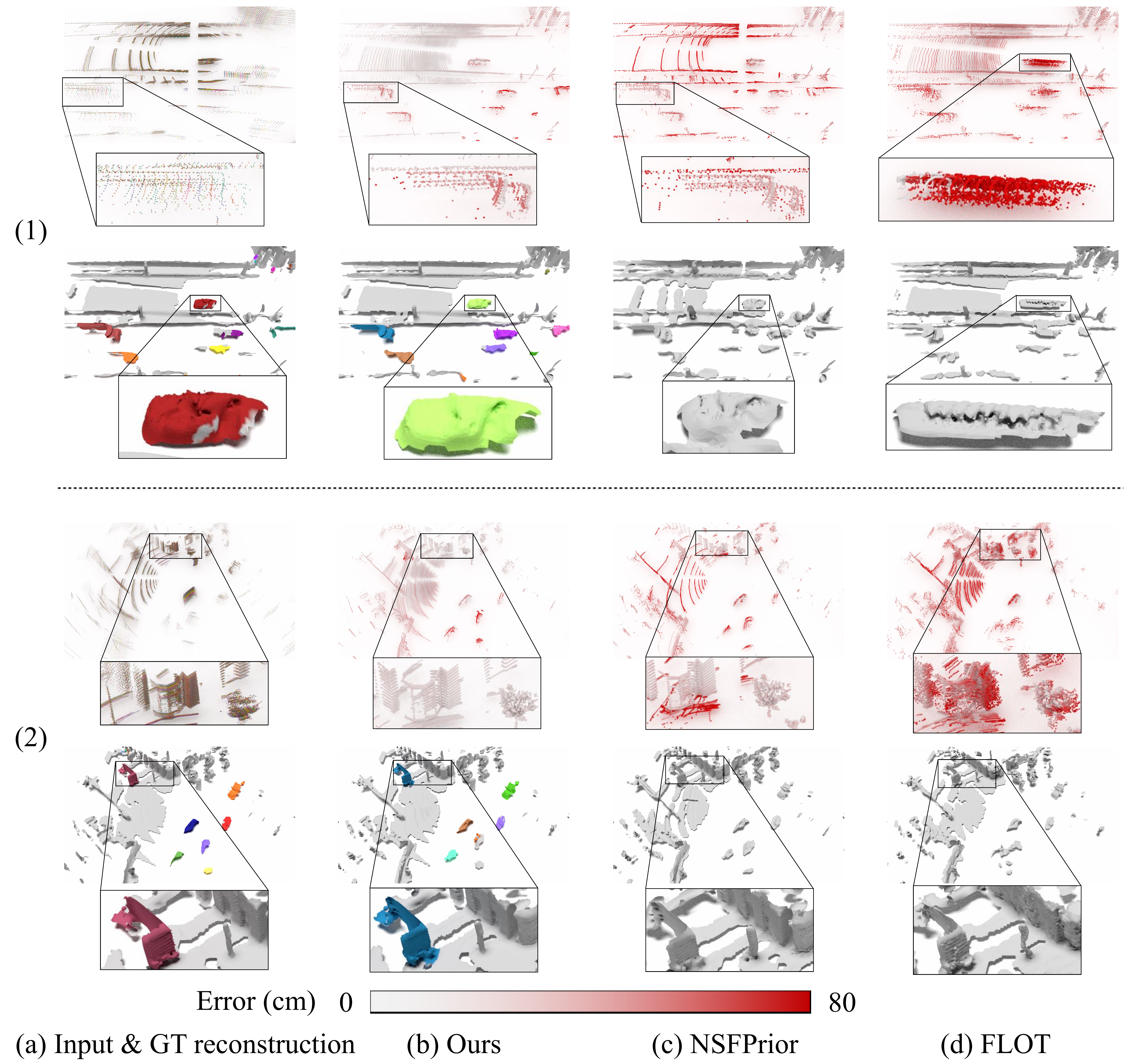}
   \vspace{-1em}
       \caption{Qualitative results showing scene flow estimation (top) and surface reconstruction (bottom) for three example scenes from the \emph{nuScenes} dataset.}
   \label{fig:supp_nuscene}
\end{figure}

We show additional qualitative results in \cref{fig:supp_waymo} and \cref{fig:supp_nuscene}. Benefiting from the explicit \textit{multi-body} assumption, our model achieves accurate scene flow estimation of both static parts (\cref{fig:supp_waymo} (1) and \cref{fig:supp_nuscene} (2)) and dynamic parts (\cref{fig:supp_waymo} (3) and \cref{fig:supp_nuscene}(1)). Errors in the automatically generated pseudo-ground truth are shown in \cref{fig:supp_nuscene}(3), in this case our model achieves more accurate flow estimation and reconstruction.  

\end{document}